\RequirePackage{snapshot}
\documentclass[10pt,journal,compsoc]{IEEEtran}

\newcommand\MYhyperrefoptions{bookmarks=true,bookmarksnumbered=true,
pdfpagemode={UseOutlines},plainpages=false,pdfpagelabels=true,
colorlinks=true,linkcolor={black},citecolor={black},urlcolor={black},
pdftitle={SurfelMeshing: Online Surfel-Based Mesh Reconstruction},pdfsubject={Computer Vision},pdfauthor={Thomas Schoeps, Torsten Sattler, Marc Pollefeys},pdfkeywords={3D Reconstruction, Depth Fusion, Triangulation, Loop Closure, Real-Time, Surfels}}\usepackage[\MYhyperrefoptions,pdftex]{hyperref}

\usepackage{capt-of}
\usepackage{graphicx}
\usepackage{overpic}
\usepackage{amsmath,amssymb}
\usepackage{xifthen}
\usepackage{multirow}
\usepackage{algorithm}
\usepackage{eqparbox}
\usepackage[noend]{algpseudocode}
\usepackage{xcolor}
\ifCLASSOPTIONcompsoc
      \usepackage[nocompress]{cite}
\else
    \usepackage{cite}
\fi
\usepackage{fancyhdr}
\usepackage{eso-pic}

\AddToShipoutPictureFG{
  \AtPageUpperLeft{%
    \raisebox{-1.7em}{%
      \begin{minipage}{\paperwidth}%
        \centering
        \tiny%
This is the author's version of an article that has been published in this journal. Changes were made to this version by the publisher prior to publication.\\
The final version of record is available at \enspace\enspace\enspace http://dx.doi.org/10.1109/TPAMI.2019.2947048%
      \end{minipage}%
    }%
  }%
  \AtPageLowerLeft{%
    \raisebox{0.5em}{%
      \begin{minipage}{\paperwidth}%
        \centering
        \tiny%
Copyright (c) 2019 IEEE. Personal use is permitted. For any other purposes, permission must be obtained from the IEEE by emailing pubs-permissions@ieee.org.
      \end{minipage}%
    }%
  }%
}

\newcommand{\PAR}[1]{\vskip4pt \noindent{\bf #1~}}

\newcommand{\COM}[1]{\hfill\#\ #1}  

\usepackage{xspace}

\makeatletter
\DeclareRobustCommand\onedot{\futurelet\@let@token\@onedot}
\def\@onedot{\ifx\@let@token.\else.\null\fi\xspace}

\def\eg{\emph{e.g}\onedot} 
\def\ie{\emph{i.e}\onedot} 
\def\cf{\emph{c.f}\onedot} 
\def\etc{\emph{etc}\onedot} 
 
\def\etal{\emph{et al}\onedot}
\makeatother

\begin{document}

\title{SurfelMeshing: Online Surfel-Based\\Mesh Reconstruction}

\author{Thomas~Sch\"ops,         Torsten~Sattler,         and~Marc~Pollefeys\IEEEcompsocitemizethanks{\IEEEcompsocthanksitem Thomas Sch\"{o}ps and Marc Pollefeys are with the Department of Computer Science, ETH Zurich, Switzerland.\protect\\
E-mail: firstname.lastname@inf.ethz.ch
\IEEEcompsocthanksitem Marc Pollefeys is additionally with Microsoft, Zurich.
\IEEEcompsocthanksitem Torsten Sattler is with Chalmers University of Technology.\protect\\
E-mail: torsat@chalmers.se}}

\markboth{IEEE TRANSACTIONS ON PATTERN ANALYSIS AND MACHINE INTELLIGENCE,~VOL.~XX, NO.~X, September~2018}{Sch\"ops \MakeLowercase{\textit{et al.}}: SurfelMeshing: Online Surfel-Based Mesh Reconstruction}

\IEEEtitleabstractindextext{\begin{abstract}
We address the problem of mesh reconstruction from live RGB-D video, assuming a calibrated camera and poses provided externally (\eg, by a SLAM system). In contrast to most existing approaches, we do not fuse depth measurements in a volume but in a dense surfel cloud. We asynchronously (re)triangulate the smoothed surfels to reconstruct a surface mesh. This novel approach enables to maintain a dense surface representation of the scene during SLAM which can quickly adapt to loop closures. This is possible by deforming the surfel cloud and asynchronously remeshing the surface where necessary. The surfel-based representation also naturally supports strongly varying scan resolution. In particular, it reconstructs colors at the input camera's resolution. Moreover, in contrast to many volumetric approaches, ours can reconstruct thin objects since objects do not need to enclose a volume. We demonstrate our approach in a number of experiments, showing that it produces reconstructions that are competitive with the state-of-the-art, and we discuss its advantages and limitations. The algorithm (excluding loop closure functionality) is available as open source at \url{https://github.com/puzzlepaint/surfelmeshing}.
\end{abstract}

}

\maketitle

\IEEEdisplaynontitleabstractindextext

\IEEEpeerreviewmaketitle

\ifCLASSOPTIONcompsoc
\IEEEraisesectionheading{\section{Introduction}\label{sec:introduction}}
\else
\section{Introduction}
\label{sec:introduction}
\fi

\begin{figure*}[t]
\begin{center}
\includegraphics[trim={0 0 0 0px},clip,width=1.0\linewidth]{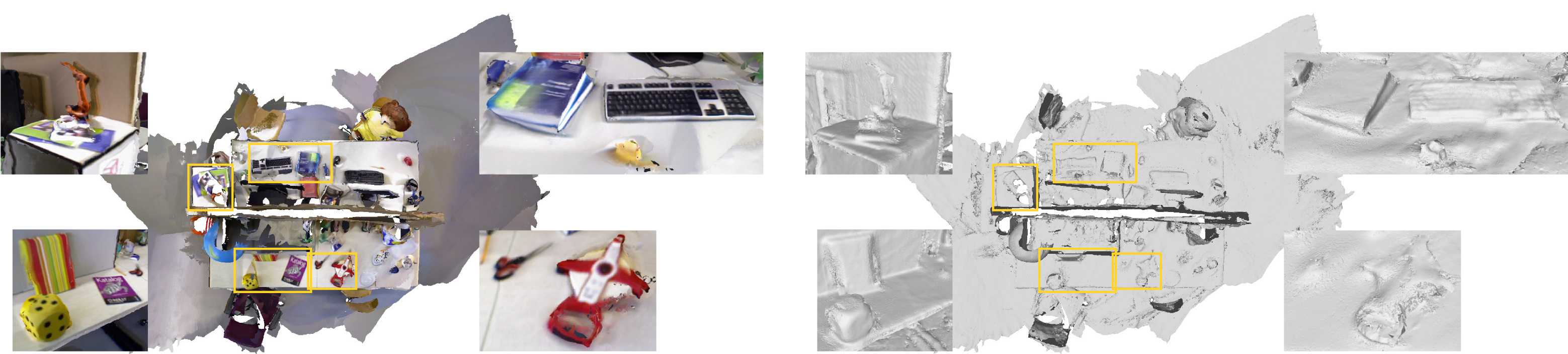}
\end{center}
\caption{Triangle mesh reconstructed with our method with loop closures handled during reconstruction, colored (left) and shaded. }
\label{fig:teaser}
\end{figure*}
\IEEEPARstart{T}{he} availability of fast programmable GPUs has enabled handling the massive amount of RGB-D data generated by depth sensors such as the Kinect in real-time. 
This has led to dense Simultaneous Localization and Mapping (SLAM) systems~\cite{newcombe2011kinectfusion,Whelan14ijrr}, online 3D scene perception approaches~\cite{Schoeps2017CVIU,Haene2017IMAVIS}, and applications such as Augmented Reality~(AR)~\cite{Newcombe2011ICCV,schoeps2017ismar} building on top of SLAM and dense mapping.

Performing dense mapping in real-time is particularly important for interactions with the environment in AR or in robotics.
For example, robots navigating through unknown scenes must map their surroundings quickly to be able to advance their path planning \cite{breitenmoser2012surface,garrido2013application,bircher2015structural}.
Complex AR applications, \eg, AR games in which users can race cars through their living room,
need more information than is provided in a single depth map, which would be sufficient for occlusion handling but does not provide any information in the occluded areas.
Consumer applications benefit from online mapping which avoids waiting times for users.
Online mapping is also useful as a live preview for offline 3D reconstructions, \eg, created with hand-held mobile devices.
Especially non-expert users benefit from this as a feedback about which parts of the scene have been reconstructed in good quality already and where additional data is needed.

A central question in the context of SLAM and online dense mapping is the choice of scene representation. 
On the one side, applications such as collision detection in physics simulations~\cite{schoeps2014ismar}, occlusion handling in AR applications~\cite{Arth2011ISMAR,Newcombe2011ICCV}, and obstacle detection and path planning for autonomous navigation \cite{breitenmoser2012surface,garrido2013application,bircher2015structural} benefit from \emph{continuous surface} representations, \eg, meshes. 
On the other side, scene representations need to be \emph{flexible} to handle spatial deformations caused by loop closure events in SLAM. 
It is important that they can be \emph{updated efficiently} if new data becomes available to allow for real-time processing. 
Ideally, a scene representation should also be \emph{adaptive} in terms of resolution: 
It should be able to model both large structures and small or thin objects without too much memory overhead.

\begin{figure}[t]
\begin{center}
\renewcommand{\tabcolsep}{0px}
\renewcommand{\arraystretch}{0}
\begin{tabular}{cccc}
\includegraphics[trim={0 0px 0 0px},clip,width=0.25\linewidth]{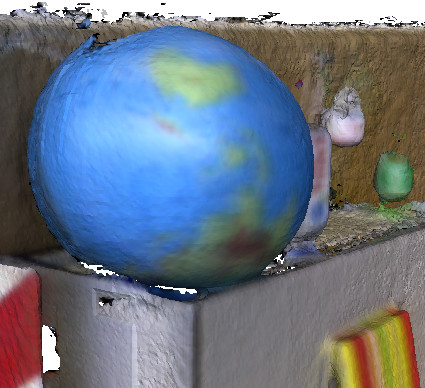}&\includegraphics[trim={0 0px 0 0px},clip,width=0.25\linewidth]{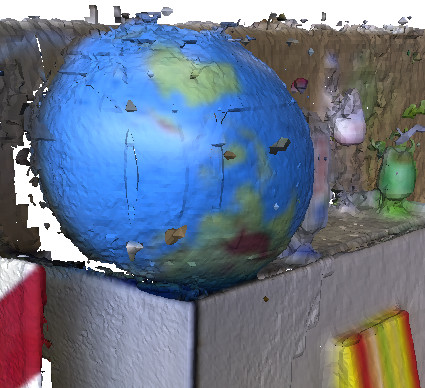}&\includegraphics[trim={0 0px 0 0px},clip,width=0.25\linewidth]{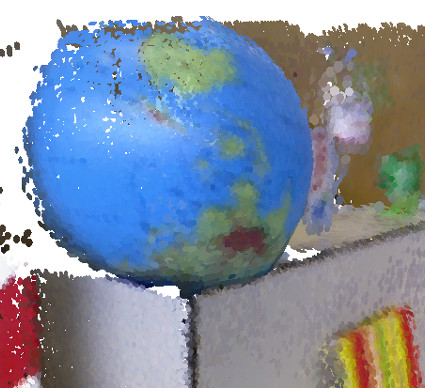}&\includegraphics[trim={0 0px 0 0px},clip,width=0.25\linewidth]{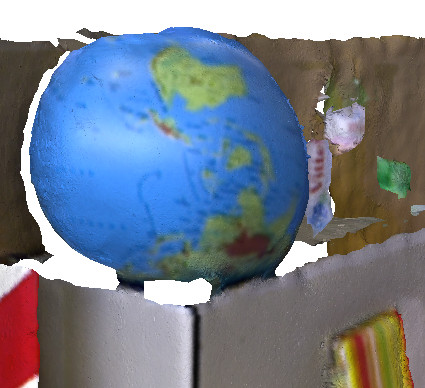}\\%
\includegraphics[trim={0 0px 0 0px},clip,width=0.25\linewidth]{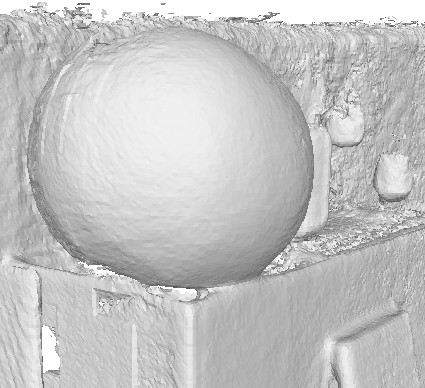}&\includegraphics[trim={0 0px 0 0px},clip,width=0.25\linewidth]{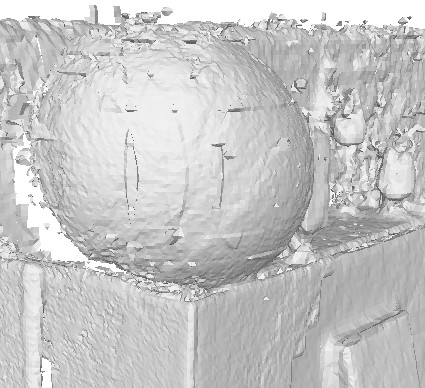}&\includegraphics[trim={0 0px 0 0px},clip,width=0.25\linewidth]{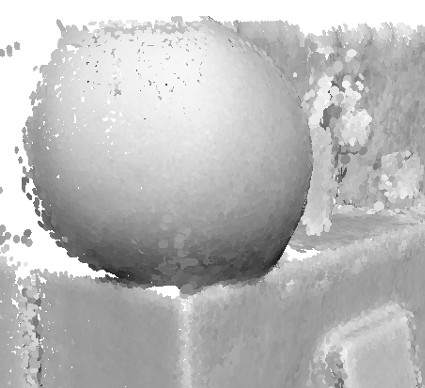}&\includegraphics[trim={0 0px 0 0px},clip,width=0.25\linewidth]{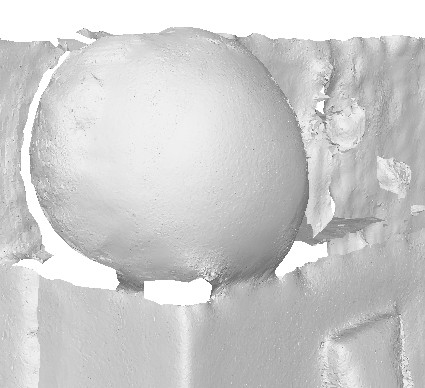}\\[0.5em]\scriptsize InfiniTAM \cite{infinitam2015ismar} & \scriptsize FastFusion \cite{steinbruecker2014volumetric} & \scriptsize ElasticFusion \cite{whelan2015elasticfusion} & \scriptsize Ours
\end{tabular}
\end{center}
\caption{Left to right: voxel-based meshes by InfiniTAM \cite{infinitam2015ismar} and FastFusion \cite{steinbruecker2014volumetric} (each with default voxel size settings), surfel splat rendering by ElasticFusion \cite{whelan2015elasticfusion}, our surfel-based mesh.
The same camera trajectory is used with all approaches.
Notice the sharper colors produced by our method. }\label{fig:representation_comparison}
\end{figure}

The predominant scene representation for RGB-D SLAM and real-time dense 3D reconstruction is a voxel volume~\cite{curless1996volumetric}. 
Each voxel in the volume stores the truncated signed distance to the closest surface, which can be updated very efficiently in parallel.
A continuous surface representation in the form of a mesh can then be extracted from this volume with the Marching Cubes algorithm \cite{lorensen1987marching}. However, volume-based methods lack flexibility: Handling loop closures during online operation is expensive, since accurate compensation can imply changing the whole volume. 
At the same time, the resolution of the voxel volume is often fixed in practice for efficiency reasons~\cite{Whelan14ijrr,kahler2016real,dai2017bundle}, which limits the adaptiveness of the representation (\cf Fig.~\ref{fig:representation_comparison}, left).
In addition, objects cannot be reconstructed if they are small or thin with respect to the voxel size.

Another popular approach is to represent the scene with a set of points or with \emph{surfels}~\cite{pfister2000surfels,whelan2015elasticfusion}, \ie, oriented discs (\cf Fig.~\ref{fig:representation_comparison}, middle right). This scene representation is flexible, since point or surfel coordinates can be updated very efficiently for the whole reconstruction. It is also highly adaptive as measurements at a higher resolution lead to denser point resp.~surfel clouds, and it easily handles thin objects. 
The main drawback of this representation is its discrete nature, which can be resolved by meshing.  
However, global methods such as Delaunay triangulation~\cite{litvinov2013incremental,romanoni2015efficient,piazza2018realtime} or Poisson reconstruction~\cite{Kazhdan2006SGP} are too computationally expensive to run in real-time on the dense point clouds generated by RGB-D sensors.
In contrast, efficient local meshing methods~\cite{gopi2000fast,marton2009fast} are very sensitive to noise, leading to both noisy surface reconstructions and holes in the meshes.

In this paper, we show that online meshing of reconstructed dense surfel clouds is possible through denoising and adaptive remeshing (\cf~Fig.~\ref{fig:teaser}). 
Our algorithm is a novel combination of surfel reconstruction \cite{keller2013real,whelan2015elasticfusion} (Sec.~\ref{sec:surfel_reconstruction}) with fast point set triangulation \cite{gopi2000fast,marton2009fast} (Sec.~\ref{sec:meshing}). 
The key step for enabling local triangulation is a novel surfel denoising scheme (Sec.~\ref{sec:denoising}) to handle the noisy input data (\cf Fig.~\ref{fig:denoising_evaluation}). 
For efficiency, flexibility, and adaptiveness, we introduce a remeshing algorithm (Sec.~\ref{sec:remeshing}) that recomputes the mesh locally if required. 
In contrast to volume-based methods, our approach easily adapts to loop closures and leads to higher resolution models (\cf Fig.~\ref{fig:representation_comparison}) while automatically handling thin objects (\cf Fig.~\ref{fig:thin_object}). 
In contrast to purely point resp.~surfel cloud-based approaches, our method reconstructs high-resolution meshes during online operation. 
To the best of our knowledge, ours is the first approach to provide a continuous surface reconstruction while allowing flexible deformations for loop closures, being adaptive to different resolutions, and allowing efficient updates.

Looking at the complete system, there is a fundamental question to
be asked about how tightly the SLAM and 3D reconstruction components
should be integrated with each other.
A tight integration allows to share data structures, allowing both
systems to influence each other while reducing duplicated efforts.
Voxel-based methods such as KinectFusion \cite{newcombe2011kinectfusion}
commonly use the volume for both tracking and mapping.
A loose integration enables to easily exchange the 3D reconstruction and
SLAM parts independently.
This improves portability, \eg, one could use ARCore resp.~ARKit
when porting a system to Android respectively iOS.
In this article, we focus on the 3D reconstruction part and present a
meshing pipeline that is rather loosely coupled, while offering the
potential for tighter integration.
In fact, making a step towards a tight integration between both parts, similar to KinectFusion, for surfel-based approaches was one of the motivations for our work.
We will briefly discuss the advantages and disadvantages of a tight
integration in the conclusion, but mainly leave this question for future
work.

\section{Related Work}
\label{sec:related_work}

This section discusses related work in the area of online triangle mesh reconstruction.
We focus on methods that create consistent scene models. We exclude offline methods as the focus of our work is on (real-time) online operation. 

\PAR{Voxel volume-based methods}
 build on truncated signed distance function (TSDF) fusion \cite{curless1996volumetric}.
Popularized by KinectFusion \cite{newcombe2011kinectfusion}, many following works focused on scaling this approach to larger scenes \cite{chen2013scalable,niessner2013TOG}, adding multi-resolution capability \cite{steinbruecker2014volumetric,kahler2016hierarchical}, or improving efficiency \cite{infinitam2015ismar,klingensmith2015chisel}. Mesh extraction typically runs at a lower frame rate compared to TSDF fusion.
In contrast to our method, these works do not handle loop closures during reconstruction.

Kintinuous \cite{Whelan14ijrr} uses a volume that moves with the camera.
Scene parts leaving this volume are triangulated~\cite{marton2009fast}, and loops are closed by deforming the mesh.
Compared to our method, \cite{Whelan14ijrr} does not merge meshes after loop closures, resulting in multiple (potentially inconsistent) meshes per surface.
K{\"a}hler \etal \cite{kahler2016real} use many small subvolumes that move independently to handle loop closures, combining 
overlapping subvolumes by weighted averaging during rendering.
Yet, subvolumes are only aligned rigidly, which can lead to inconsistencies.
BundleFusion \cite{dai2017bundle} addresses loop closures by de- and re-integrating old RGB-D frames when they move, enabling it 
to correct all drift. 
It is computationally expensive, requiring two GPUs for real-time operation, and does not scale well as it has to keep all frames in memory.  Using keyframes reduces the computational demand~\cite{maier2017efficient,han2018flashfusion}.  However, data can be lost if it cannot be integrated into a keyframe and the reconstruction quality can degrade if data is fused with keyframes early.
Our method is more efficient than \cite{dai2017bundle} while still using all the data. In addition, it handles varying scan resolution naturally and can reconstruct thin objects.

\PAR{Delaunay tetrahedralization}
of a point cloud is a way to discretize 3D space. Tetrahedra can be classified as 'inside' or 'outside' of objects and the surface can be extracted as the interface between these classes.
\cite{litvinov2013incremental} uses this approach while being able to incrementally add points to the model. \cite{romanoni2015efficient,piazza2018realtime} extend \cite{litvinov2013incremental} to allow for moving existing points.
For performance reasons, these methods are typically only applied to sparse (SLAM) point clouds. Only \cite{piazza2018realtime} runs in real-time on those. In contrast to our method, these methods thus cannot handle dense surfel clouds in real-time.

\PAR{Surfel-based methods}
represent the scene as a set of surfels \cite{pfister2000surfels}.
MRSMap \cite{stuckler2014multi} stores multi-resolution surfel maps in octrees.
It handles loop closures via pose graph optimization and generates a consistent map only afterwards. Weise \etal \cite{weise2011online} present a system for scanning small objects. Loop closures are handled in real-time by deforming the surfel cloud using a sparse deformation graph.
Another line of work \cite{keller2013real,lefloch2015anisotropic,lefloch2017comprehensive} addresses high-quality surfel-based tracking and mapping, but does not address the loop closure problem.
ElasticFusion \cite{whelan2015elasticfusion} extends \cite{keller2013real} with real-time loop closure handling similar to \cite{weise2011online}.  Gao and Tedrake \cite{gao2018surfelwarp} present a surfel-based approach for reconstructing dynamic objects, but note the lack of real-time mesh reconstruction as a drawback.
Yan \etal \cite{yan2017dense} propose a probabilistic surfel map representation for SLAM.
They reconstruct a mesh from deformed keyframe depth maps as a post-processing step. Each depth map is handled independently, potentially resulting in multiple meshes per surface. In contrast to these previous works, our approach is able to construct meshes during online operation.

\begin{figure}[t]
\begin{center}
\includegraphics[width=1.0\linewidth]{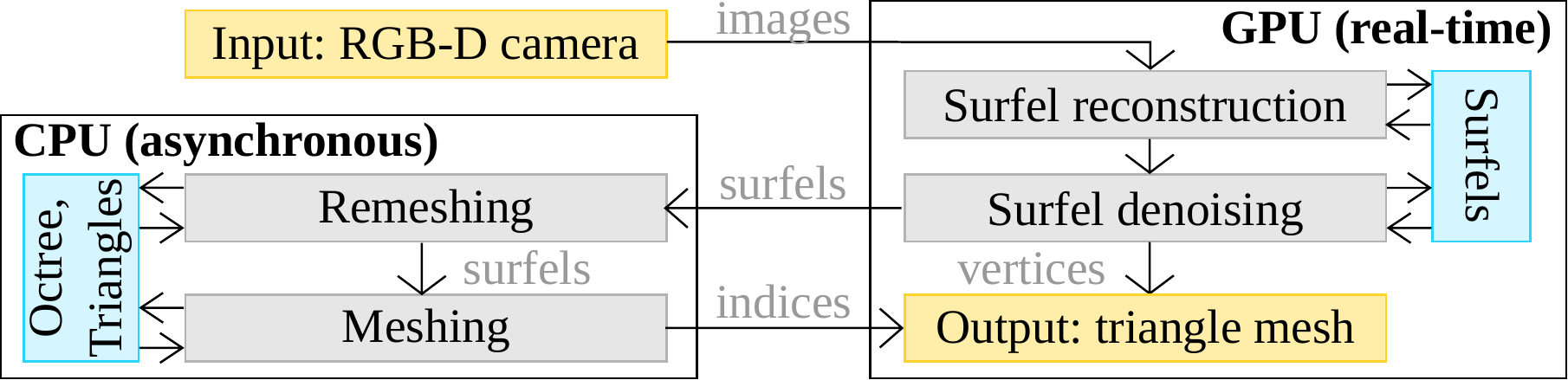}
\end{center}
\caption{Data flow overview for our algorithm.}
\label{fig:system_overview}
\end{figure}

\PAR{Meshing approaches.}
Bodenm\"uller \cite{bodenmueller2009streaming} presents a streaming mesh reconstruction method for point clouds, but does not address noisy consumer depth cameras or loop closures.
Schertler \etal~\cite{schertler2017field} incrementally reconstruct high-quality meshes from point clouds.
In contrast to our approach, their method is not suitable for real-time use on high-frequency input such as RGB-D video. 
Ladick\'y \etal \cite{ladicky2017point} learn to convert point clouds to SDF volumes for meshing using Marching Cubes. Using the same GPU as used in this work, their approach handles about 300,000 points in real-time. 
For comparison, each of our input depth map already contains this number of points. In contrast to ours, \cite{ladicky2017point} uses a fixed voxel size, limiting the mesh resolution. Zienkiewicz \etal \cite{zienkiewicz2016monocular} fit a pre-defined triangular mesh to the observations.
They explicitly handle varying observation scale but only demonstrate their method on 2.5D height maps.
In contrast, our approach handles both varying scale and general scenes.

\section{Surfel-Based Mesh Reconstruction}

The input to our algorithm is an RGB-D video stream from a calibrated camera and camera poses provided by a SLAM system such as~\cite{whelan2015elasticfusion}. 
Our algorithm consists of four main parts (\cf Fig.~\ref{fig:system_overview}): surfel reconstruction (adapted from \cite{keller2013real,whelan2015elasticfusion}), surfel denoising (new contribution), meshing (adapted from \cite{gopi2000fast,marton2009fast}), and remeshing (new contribution).
Surfel reconstruction and denoising reconstruct a surfel cloud   at the frame rate of the video input. Meshing and remeshing create and maintain a triangulation of it. The output is a mesh whose vertices correspond to the surfels, which are updated at frame rate, and whose topology is determined by the triangulation, which is updated at a lower rate.

Our approach allows for \emph{efficient} integration of new video frames by running the expensive tasks of meshing and remeshing asynchronously in the background, similar to asynchronous mesh extraction in volume-based methods~\cite{klingensmith2015chisel}.
\emph{Flexibility} and \emph{adaptiveness} are given by the surfel representation, which can be quickly deformed to adapt to loop closures and can model scenes of varying resolution.
In addition, the meshing approach needs to preserve these advantages: For example, meshing with fixed-resolution volumes would lose adaptiveness.
Our approach thus exploits that deforming the surfels directly deforms the mesh as well without additional effort (flexibility), and operates at the resolution of the surfel cloud (adaptiveness).
We describe each component of our approach in detail below. 

\subsection{Surfel Reconstruction}
\label{sec:surfel_reconstruction}
Our algorithm triangulates a surfel cloud. 
To reconstruct this cloud, we slightly adapt  \cite{keller2013real,whelan2015elasticfusion}. The following outlines the relevant parts of this approach including our modifications. Sec.~\ref{sec:denoising} presents our extensions for surfel denoising. 

\begin{figure}[t]
\begin{center}
\includegraphics[width=0.999\linewidth]{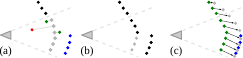}
\end{center}
\caption{Illustration of surfel reconstruction: (a) Data association. New measurements (gray squares) are associated with existing surfels as follows: Surfels outside the camera view are not considered (black dots). Green surfels are \emph{supported} by the nearby measurements. The red surfel \emph{conflicts} with the measurement behind it. The blue surfels are \emph{occluded} by the measurements. (b) After integrating the measurements from (a), the conflicting surfel has been replaced, the supported surfels have been averaged with the measurements, and new surfels (gray) have been created for measurements without a supported or conflicting surfel. (c) Loops are closed with a non-rigid deformation, aiming to align new surface parts (green) with corresponding old surfaces (blue).
}
\label{fig:surfel_reconstruction_sketch}
\end{figure}

From the input stream of RGB-D images, we construct and maintain an unordered set of surfels covering the visible surfaces.
Once a new frame becomes available, a data association step decides for each new depth measurement, \ie, each pixel having a depth value, whether it creates a new surfel or is used to refine existing surfels. 
In the event of a loop closure, the surfel cloud is deformed to align matching surface parts. See also Fig.~\ref{fig:surfel_reconstruction_sketch} for an overview of this process.

\PAR{Surfel representation.}
Each surfel $s$ comprises a 3D position $\mathbf{p}_s$, a normal $\mathbf{n}_s$, an RGB color $\mathbf{c}_s$, a confidence score $\sigma_s$, a radius $r_s$ used for neighbor search, a creation timestamp $t_{s,0}$, and a latest update timestamp $t_s$.

\PAR{Data association.}
Similar to \cite{keller2013real}, we project the surfels into each new RGB-D image to determine which depth measurements should be associated with existing surfels.
To determine which surfel projects to which pixel, \cite{keller2013real} creates a super-sampled index map rendering of all surfels, which is however still limited in resolution.
We improve upon this by always taking surfel indices directly from the result of the projection operation, thus avoiding to store surfel indices in a map with limited space.
The projection is repeated in each step where these indices are required.

For data association, each surfel is tested against the pixel it projects to, as well as the neighboring pixel that is closest to the projected sub-pixel position, and can potentially be associated to both.
We use a simple model of the measurement uncertainty which defines the uncertainty range of a measurement with depth $z$ as the depth interval $[(1 - \gamma) z, (1 + \gamma) z]$, with $\gamma = 0.05$.
Both the sensor depth uncertainty and small pose uncertainties should be accounted for here.
We found that this simple model worked well in the scenarios tested, however, more sophisticated models (which ideally tightly model the uncertainty of poses and 3D structure estimated by the underlying SLAM system) could be easily substituted. 

Based on this model, we classify each surfel to be either \emph{conflicting} with a depth measurement it projects to, to be \emph{occluded} by it, or \emph{supported} by it.
A surfel \emph{conflicts} with the measurement if it projects in front of the measurement uncertainty range, thus violating the constraint that the space between the measurement and the camera must be free. 
A surfel is \emph{occluded} by the measurement if it is not conflicting, and either projects behind the uncertainty depth range, or its normal points away from the camera or differs too strongly from the measurement normal.
The remaining surfels are \emph{supported by} the measurement.

\PAR{Measurement integration.}
As in~\cite{keller2013real}, we create a new surfel for each pixel that is not associated with any conflicting or supported surfel.
Surfels are thus created to match the images' resolutions, which makes the approach adaptive to the input resolution.
For a new surfel initialized from a pixel, we set the initial surfel position $\mathbf{p}_s$ to the un-projected position of the pixel and estimate its normal $\mathbf{n}_s$ via finite differences in the depth image.
The pixel color is used as the surfel color $\mathbf{c}_s$, the confidence $\sigma_s$ is set to 1, and both timestamps are set to the current timestamp.
In our method, the surfel radius $r_s$ represents the expected maximum distance to neighboring surfels on the same surface. 
That is, $r_s$ is the maximum distance of the surfel to all surfels it should later be connected to during meshing.
For a depth measurement at pixel $(x, y)$, we compute $r_s$ as \begin{equation}
r_s = 1.5 \cdot \max_{u, v \in \{-1, 0, 1\}} ||p(x, y) - p(x + u, y + v)||_2 \enspace ,
\label{eq:radius}
\end{equation}
where $p(x,y)$ is the 3D point corresponding to pixel $(x, y)$. Intuitively, we chose the radius $r_s$ to contain the direct neighbors in image space, providing us with an estimate of how far away neighboring surfels can be in 3D space. 
To avoid noisy radii, we only use pixels for which all neighbors in the 8-neighborhood have valid depth measurements.

Integration of a measurement into a surfel is performed as in \cite{keller2013real},
with the difference that 
we integrate a measurement into \emph{all} surfels that are supported by it to obtain an as-smooth-as-possible reconstruction.
For a given measurement $m$ and surfel $s$, we compute the weighted averages of $\mathbf{p}_s$, $\mathbf{n}_s$, and $\mathbf{c}_s$ with the measured values $\mathbf{p}_m$, $\mathbf{n}_m$, and $\mathbf{c}_m$, and update the timestamp $t_s$ and confidence $\sigma_s$:
\begin{equation}
\text{With} ~ f(\mathbf{v}_s, \mathbf{v}_m) = \frac{\sigma_s \mathbf{v}_s + w \mathbf{v}_m}{\sigma_s + w}
 : \!
\begin{array}{c}
\mathbf{p}_s := f(\mathbf{p}_s, \mathbf{p}_m),\\
\mathbf{\hat{n}}_s := f(\mathbf{n}_s, \mathbf{n}_m),\\
\mathbf{c}_s := f(\mathbf{c}_s, \mathbf{c}_m).
\end{array} \hspace{-0.4em}
\end{equation}
\noindent\begin{minipage}{.35\linewidth}
\begin{equation}
  \mathbf{n}_s := \frac{\mathbf{\hat{n}}_s}{| \mathbf{n}_s |}
\end{equation}
\end{minipage}\begin{minipage}{.65\linewidth}
\begin{equation}
  \sigma_s := \min\{ \sigma_s + w, \sigma_{\text{max}} \}
\end{equation}
\end{minipage}\\[0.7em]
The weight $w$ is computed as $\frac{1}{|S_m|}$, with $S_m$ being the set of all surfels supported by the measurement.
We clamp the surfel confidence at a low maximum value $\sigma_{\text{max}} = 5$ such that the weight of new measurements stays high.
This allows existing surfaces and new surfels within them to quickly converge to similar attribute values.
If computing Eq.~\ref{eq:radius} on the current image leads to a smaller radius, we update the radius $r_s$ to this smaller value. 
Nearby surfels having very similar attributes are merged as in \cite{keller2013real}.

We decrease the confidence of conflicting surfels by one.
Once the confidence of a conflicting surfel reaches zero, it is replaced with the new measurement it conflicts with, as described above for the creation of new surfels.

\PAR{Loop closures.}
When a loop is closed by the SLAM system, the surfels need to be adapted to the SLAM system's updated state estimate.
It should be ensured that the change to the estimated geometry is performed consistently in the reconstruction and in the SLAM system.
Since we use ElasticFusion \cite{whelan2015elasticfusion} as SLAM system, we employ its loop closure handling procedure.
We use an active window and only consider \emph{active} surfels whose last update timestamp $t_s$ is not too old for data association and integration.
In the event of a loop closure, the surfel cloud is deformed based on the surfel timestamps and a new position and normal is computed for each surfel.
Surfels outside the active window that are consistent with active surfels after the loop closure are re-activated.
We refer to \cite{whelan2015elasticfusion} for details.
Concerning meshing, surfel movements due to loop closures are handled exactly like surfel movements due to integrating new measurements, with the exception of a small performance optimization mentioned in Sec.~\ref{sec:remeshing}.
This makes the procedure to handle loop closures very efficient and allows our approach to provide a flexible surface representation.

\subsection{Surfel Denoising}
\label{sec:denoising}

There are various sources of noise such as depth camera noise, camera calibration, and pose inaccuracies.
Thus, independently reconstructed surfels are usually noisy, despite the averaging used in measurement integration.
This in particular applies to new surfels, causing problems as triangulation is sensitive to noise.
We thus introduce a regularization step.
Furthermore, discontinuities are likely to arise at the boundaries of the integrated depth images. We address them with a blending step.
This section presents these two contributions.

\begin{figure*}[t]
\begin{center}
\includegraphics[trim={0 0 0 0},clip,height=6em]{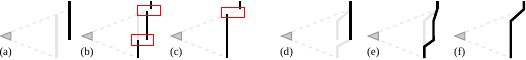}
\end{center}
\caption{Illustration of raw depth map integration, (a) - (c), vs.~observation boundary blending, (d) - (f).
(a) The camera observes an existing surface (black) in a different position (gray), \eg, due to slight pose drift.
(b) After integrating one new observation of the gray surface, 
discontinuities (red) are created. (c) After integrating many new observations, 
there is still a discontinuity, causing a hole in the mesh.
(d) Observation hallucinated by Alg.~\ref{alg:observation_boundary_blending} based on (a).
(e) After integrating one new observation of the gray surface, 
no discontinuities are created.
(f) After integrating many new observations, blending still avoids discontinuities.
In contrast, using a lower integration weight for boundary regions would lead to situation (c).
}
\label{fig:boundary_blending}
\end{figure*}

\PAR{Regularization.}
Regularization aims to maintain a smoothed version of the surfel cloud, reducing noise among neighboring surfels.
We therefore extend the surfel structure with a denoised position $\mathbf{\bar{p}}_s$ (initialized to $\mathbf{p}_s$), indices $N_s$ of four neighboring surfels, and temporary storage for accumulating cost gradients.  Our proposed regularization works across transitive neighbors: each surfel's denoised position depends on the denoised positions of its neighbors, which again depend on the neighbors' neighbors, and so on.
Thus, 
we do not require $N_s$ to contain the closest neighbors. 
This enables us to use a simple neighborhood selection scheme suitable for GPU processing: 
During data association (Sec.~\ref{sec:surfel_reconstruction}), we create an index image which for each pixel references one random supported surfel (to keep computation time low).
Due to the transitivity, this
random choice is not significant. Afterwards, for each supported surfel $s$, we examine the 4-neighborhood of the pixel it projects to. We update $N_s$ to contain the up to four closest neighbors from both the surfel's existing neighbors $N_s$ and the supported surfels assigned to the pixels in the 4-neighborhood. 
Neighbors farther away from $\mathbf{p}_s$ than $2\cdot r_s$ are not considered. 

Our denoising approach then optimizes a cost $C$, which contains a residual for each surfel $s$ in the surfel cloud $S$:
\begin{equation}
C(S) = \sum_{s \in S} r_{\text{data}}(s) + w_{\text{reg}} r_{\text{reg}}(s) \enspace .
\end{equation}
The cost consists of a data term $r_{\text{data}}$ and regularizer $r_{\text{reg}}$.
We use a weighting factor $w_{\text{reg}}$ to weigh the two terms.
It depends on the amount of noise caused by uncertainties in the camera measurements and calibration, pose estimates, \etc, and we empirically set it to 10 to prioritize smoothness.  The data term keeps the denoised surfel position $\mathbf{\bar{p}}_s$ close to the position $\mathbf{p}_s$ where it was measured:
\begin{equation}
r_{\text{data}}(s) = ||\mathbf{\bar{p}}_s - \mathbf{p}_s||_2^2 \enspace .
\end{equation}
The regularizer moves the surfel close to its neighbors $N_s$, measured along the surfel's normal direction $\mathbf{n}_s$:
\begin{equation}
r_{\text{reg}}(s) = \frac{1}{|N_s|}
                    \sum_{n \in N_s} \left( \mathbf{n}_s^T (\mathbf{\bar{p}}_n - \mathbf{\bar{p}}_s) \right)^2  \enspace .
\end{equation}
Notice that this formulation is different from basic Laplacian smoothing in that it applies a clearly defined smoothing strength (controlled by $w_{\text{reg}}$) to the surfels upon convergence, and it also takes the surfel normals into account.

After each iteration of surfel reconstruction, we run one iteration of gradient descent to minimize the cost $C$.
We choose the descent step length individually for each surfel $s$ as follows, which is empirically normalized to a relative length of $0.5$ to obtain good convergence:
\begin{equation}
\label{eq:stepsize}
0.5 \cdot \bigg(
  1 +
  w_{\text{reg}} +
  \sum_{i \in S | s \in N_i} \frac{w_{\text{reg}}}{|N_i|}
\bigg)^{-1} \enspace .
\end{equation}
This step size normalizes the surfel's gradient length by the total residual weight it receives.
The data term contributes weight 1, the regularization term contributes weight $w_{\text{reg}}$, and the third term in the sum arises from the surfel acting as a neighbor $n$ in the regularization term of other surfels.

For performance reasons, we use an active window for denoising: We only update surfels whose latest update timestamp is within the last 30 frames.
Other surfels 
usually do not move significantly anymore since they have been unobserved for a while and thus nearly converged.

After an optimization iteration, we update the vertices of the output mesh by assigning the new surfel positions to them.
For surfels replaced by a conflicting measurement since the last triangulation iteration, we remove all incident triangles until the next meshing update to avoid artifacts.

We also use the surfel neighbors for smoother deformations on loop closures:
Instead of directly adding the computed position offset $\mathbf{\delta p}_s$ to $\mathbf{\bar{p}}_s$ and $\mathbf{p}_s$, we first average it among the neighbors for 100 iterations (chosen to be large while not causing a too strong performance hit): $\mathbf{\delta p}_s := \frac{1}{|N_s|} \sum_{n \in N_s} \mathbf{\delta p}_n$.
This yields more consistent offsets among neighbors, thus surfaces are less likely to rip apart.

\PAR{Observation boundary blending.}
Boundaries between the observed and currently unobserved parts of the scene are likely to cause surface discontinuities. 
The observed parts will be updated to match any depth bias which is currently present (\eg due to drift) while the unobserved parts remain constant (\cf Fig.~\ref{fig:boundary_blending} (a) - (c)).
This is a general issue arising from noisy input and affects all 3D scene representations: 
For volume-based methods it creates step artifacts in the reconstructed surface. 
For surfel-based methods, it additionally creates holes in the reconstruction.
Thus, we address this to reduce undesirable hole artifacts.
We hallucinate a smooth transition from the measured depth to the surfel depth at observation boundaries (determined from the data association result from Sec.~\ref{sec:surfel_reconstruction}).
This is similar to reducing the integration weight at boundaries, but in contrast to that avoids drift if the biased observations occur over many frames (\cf Fig.~\ref{fig:boundary_blending} (d) - (f) for an illustration).

The exact procedure for boundary blending is specified in Alg.~\ref{alg:observation_boundary_blending}.
This image-based algorithm first detects observation boundaries in the depth image by finding pixels which both have a depth measurement and an associated supported surfel, as well as at least one neighbor pixel which lacks at least one of these.
These pixels are used as seeds: The difference between the measurement depth and supported surfel depth at these pixels is computed and dilated to suitable neighbor pixels (which are on the side of the boundary that will be corrected) for a number of iterations.
Each pixel which receives a propagated depth difference is then updated: We hallucinate a change to the pixel's depth to reduce the depth difference, while linearly decreasing this effect the farther a pixel is away from the observation boundary.
This creates smooth depth transitions, which can prevent discontinuous step artifacts from being created.

\algrenewcommand\algorithmicindent{1.0em}\begin{algorithm}[t]
\caption{Observation boundary blending}\label{alg:observation_boundary_blending}
\begin{algorithmic}[1]
\State \# $D(p)$ : depth of pixel $p$; is set to 0 if there is no depth measurement there
\State \# $\text{SD}(p)$ : average supported surfel depth of pixel $p$; is set to 0 if there is no supported surfel at this pixel
\State \# $I_d, I_s, \delta_d, \delta_s$ : temporary image-sized buffers
\State \# $i_\text{count} = 10$ : iteration count

\Procedure{Blend}{}
\State \# Initialize $I_d$, $I_s$ to -1
\ForAll{pixels $p$}
  $I_d(p) := -1;~I_s(p) := -1$
\EndFor
\State \# For all pixels with depth and surfel(s) ...
\ForAll{pixels $p$ with $D(p) \neq 0$ and $\text{SD}(p) \neq 0$}   \State $\text{d} := \text{SD}(p) - D(p)$ \COM{Surfel-vs-measurement delta}
  \ForAll{pixels $q$ in $p$'s 8-neighborhood}
  	\State \# At boundary of measurement area?
    \If{$I_d(p) = -1 \text{ and } D(q) = 0$}       \State \# Start propagation (1st case)
      \State $\delta_d(p) := \text{d};~I_d(p) := 0;~D(p) := \text{SD}(p)$     \EndIf
    \State \# At boundary of surfel area?
    \If{$I_s(p) = -1 \text{ and } \text{SD}(q) = 0$}       \State \# Start propagation (2nd case)
      \State $\delta_s(p) := \text{d};~I_s(p) := 0$     \EndIf
  \EndFor
\EndFor
\For{$i \in [1, i_\text{count} - 1]$} \COM{Perform blending iterations}
  \State \# For all pixels with depth ...
  \ForAll{pixels $p$ with $D(p) \neq 0$}     \State \# Propagate among pixels with surfel(s)?
    \If{$\text{SD}(p) \neq 0$ and $I_d(p) = -1$}       \State \Call{Update}{$i, p, \delta_d, I_d$} \COM{Update (1st case)}
    \EndIf
    \State \# Propagate among pixels without surfels?
    \If{$\text{SD}(p) = 0$ and $I_s(p) = -1$}       \State \Call{Update}{$i, p, \delta_s, I_s$} \COM{Update (2nd case)}
    \EndIf
  \EndFor
\EndFor
\EndProcedure

\Function{Update}{$i, p, \delta, I$}
  \State \# Average deltas of previous iteration ...
  \State $\text{sum} := 0;~\text{count} := 0$   \ForAll{pixels $q$ in $p$'s 8-neighborhood}     \If{$I(q) = i - 1$}
      $\text{sum} \text{ += } \delta(q);~$count += 1
    \EndIf
  \EndFor
  \State \# Apply the (weighted) averaged delta?
  \If{count $>$ 0}     \State $I(p) := i;~\delta(p) := \frac{\text{sum}}{\text{count}};~$
    $D(p) \text{ += } (1 - \frac{i}{i_\text{count}}) \frac{\text{sum}}{\text{count}}$
  \EndIf
\EndFunction
\end{algorithmic}
\end{algorithm}

\begin{figure*}[t]
\begin{center}
\includegraphics[width=0.999\linewidth]{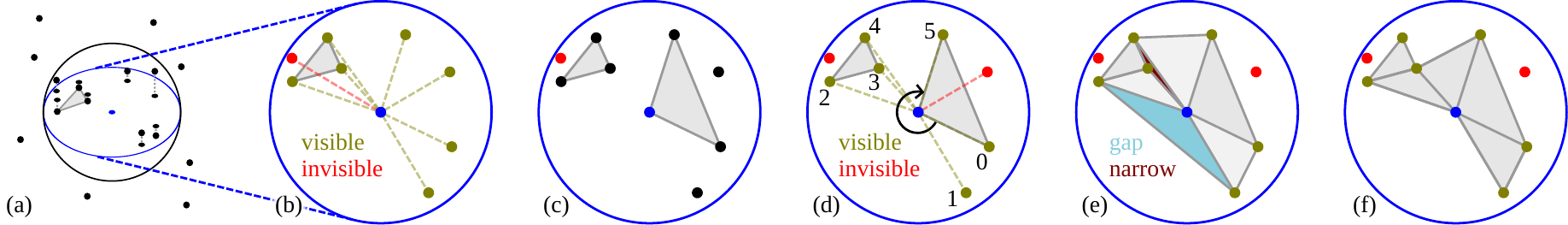}
\end{center}
\caption{Overview of triangulation for the center surfel (following \cite{gopi2000fast}): (a) Neighbor search and projection onto the tangent plane of the surfel, shown in (b) with visibility in the plane. (c) After initial triangle creation. (d) Updated visibility and neighbor ordering. (e) Gap and narrow triangle classification. (f) Result after gap and narrow triangle removal.}
\label{fig:triangulation_sketch}
\end{figure*}

\subsection{Meshing}
\label{sec:meshing}
We require a very fast, scale-independent (re)meshing algorithm to maintain a triangulation of millions of surfels during reconstruction.
For meshing, we adapt \cite{gopi2000fast,marton2009fast} with minor modifications.
We briefly present this approach with our modifications here. 
\PAR{Spatial access.}
The meshing algorithm needs to quickly and accurately find all other surfels within a surfel's radius.
Thus, a spatial access structure is needed.
For our algorithm, this structure must also efficiently adapt to moved surfels. 
In contrast to \cite{gopi2000fast,marton2009fast}, which assume that the 3D surfel positions are static, we thus use a compressed octree \cite{clarkson1983fast} which we update lazily: A moved surfel is only propagated up in the tree to the first node which contains its new position.
We only propagate the surfel downwards when a search traverses that node,  creating new nodes if required. 
Propagation stops if a leaf node or a node not traversed by the search is reached. The octree provides scale-independence with an asymptotic search time of $\mathcal{O}(n)$. While this worst-case hardly occurs in practice,
voxel hashing~\cite{niessner2013TOG} might be a faster alternative if the scan resolution is fixed and known, in particular since an efficient parallel implementation of the lazy octree might be hard. 

\PAR{Triangulation.}
The meshing algorithm greedily iterates over all surfels, and for each one locally triangulates the surfels in its neighborhood to incrementally grow the mesh.
Similar to~\cite{gopi2000fast}, each surfel is assigned a triangulation state: 
\emph{free} if the surfel has no incident triangles, \emph{front} (which corresponds to \emph{fringe} and \emph{boundary} in \cite{gopi2000fast}) if it lies on the current boundary of the mesh, or \emph{completed} if it lies within the mesh.
All new surfels are initially marked \emph{free}.
The algorithm then greedily iterates over all new surfels and all \emph{free} and \emph{front} surfels which moved since the last triangulation iteration to mesh them.
We also iterate over all surfels queued for (re)meshing (\cf~Sec.~\ref{sec:remeshing}). 

For the current surfel $s$ in this loop, the algorithm from \cite{gopi2000fast} first finds all neighboring surfels as candidates for triangulation.
In our implementation, we use the surfel's radius $r_s$ as search radius (\cf Fig.~\ref{fig:triangulation_sketch}(a)). 
$r_s$ is defined based on the 3D distance to the surfels that are neighbors to $s$ under the highest resolution under which $s$ was observed (\cf Eq.~\ref{eq:radius}). 
Thus, surfels farther away than $r_s$ can be ignored safely. 
Furthermore, if $s$ is on the mesh boundary and its boundary neighbors are further away than $r_s$, we extend the search radius as necessary to include those neighbors, up to $2\cdot r_s$.
This helps in meshing surfaces observed under slanted angles without having to resample these surfaces as done in \cite{marton2009fast}.
If this limit is exceeded, triangulation for this surfel is aborted. 

All neighbor surfels are projected onto the tangent plane defined by the surfel's normal $\mathbf{n}_s$ (\cf Fig.~\ref{fig:triangulation_sketch}(a,b)).
Neighboring surfels which are invisible from $s$ as seen in this 2D projection (\cf \cite{gopi2000fast} for details and Fig.~\ref{fig:triangulation_sketch}(b,d) for an illustration), or whose normal differs too much from $\mathbf{n}_s$, are discarded. If $s$ is \emph{free}, the algorithm from \cite{marton2009fast} first attempts to create an initial triangle with $s$ as one of its vertices (\cf Fig.~\ref{fig:triangulation_sketch}(c)).
The remaining visible neighbors are sorted according to their angle to $s$ in the 2D projection (\cf Fig.~\ref{fig:triangulation_sketch}(d)).
Spaces between adjacent neighbors are classified as \emph{gap} resp.~\emph{narrow} if the angle between them is considered too large resp.~small for triangulation (\cf Fig.~\ref{fig:triangulation_sketch}(e)).
Unless this leads to holes, neighbors which would form narrow triangles are discarded to avoid degenerate triangles~\cite{gopi2000fast}. The exact thresholds are not critical to our algorithm.
We remove the \emph{gap} classification where this closes a hole in the mesh.
Finally, new triangles are added to fill the space between $s$ and the remaining neighbors, excluding \emph{gaps} (\cf Fig.~\ref{fig:triangulation_sketch}(f)).
After a meshing iteration finishes, we update the triangle indices of our mesh to the new triangulation.

\subsection{Remeshing}
\label{sec:remeshing}
\begin{figure}[t]
\begin{center}
\includegraphics[width=0.999\linewidth]{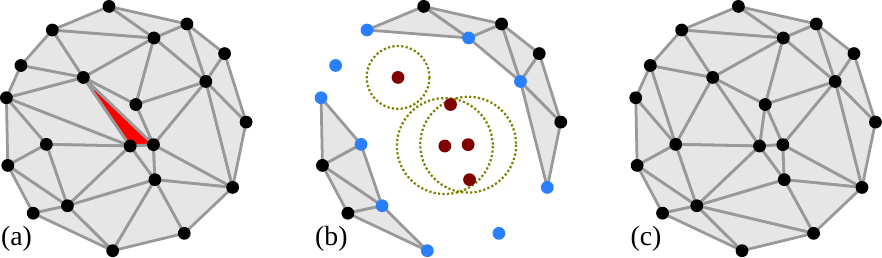}
\end{center}
\caption{Overview of the remeshing process: (a) Invalid triangles (red) are identified (\cf Sec.~\ref{sec:remeshing}). (b) All triangles connected to surfels (red) within the neighbor search radii (yellow) of the triangles' corner vertices are deleted. Affected surfels (red and blue) are scheduled for remeshing. (c) Holes are filled by the meshing algorithm (Sec.~\ref{sec:meshing} / Fig.~\ref{fig:triangulation_sketch}). }
\label{fig:remeshing_sketch}
\end{figure}
When surfels move or new surfels are created, existing parts of the mesh can become outdated and need to be updated efficiently.
For this we propose a remeshing approach, which works on the existing mesh (with vertex positions updated from the current surfel positions) before each meshing iteration.
The remeshing approach complements the meshing algorithm of Sec.~\ref{sec:meshing}: We define a triangle to be \emph{invalid} if it is impossible for the meshing algorithm to generate it given the current surfel cloud. 
We then identify \emph{invalid} triangles in the existing mesh and locally re-create the surface there using the meshing algorithm.
This is sketched in Fig.~\ref{fig:remeshing_sketch}.

A triangle can become invalid when, due to new input, surfels move or new surfels are created.
However, most triangles usually stay valid  if the surfels do not move too much. We derive the following criteria to identify valid triangles from the algorithm from Sec.~\ref{sec:meshing}:  1) For a valid triangle, at least one of its vertex surfels (named $s$ in the following) contains all triangle vertices in its neighbor search radius and has a similar normal to them (otherwise the triangle vertices could not be found during neighbor search).
2) The triangle normal must be within $90^{\circ}$ of surfel $s$' normal $\mathbf{n}_s$ (otherwise the triangle would be upside down in $s$' tangent plane).
3) In $s$' tangent plane, no other neighbor surfel projects into the triangle and the triangle does not intersect another triangle (otherwise vertex visibility would prevent the triangle's creation). 

Since gaps can be filled for hole closing and some narrow triangles cannot be removed, these are not criteria for invalid triangles.
For efficiency reasons, we modify the criteria as follows: 
a) We allow triangles to become 1.5 times longer than the maximum extended neighbor search radius in order to allow for more surfel movement before remeshing. b) We do not fully test for criterion 3) to avoid performing many (slow) spatial searches.  As a frequent special case of this criterion, we address the case of newly created surfels next to existing triangles:
we find all existing surfels within the neighbor search radius of new surfels and delete all triangles connected to them.
However, we do not test for conflicts arising for existing surfels or triangles.
We found that this test is not strictly required, as affected regions typically are remeshed anyways due to violating other criteria.

In our remeshing algorithm, we test all triangles whose vertices have moved since the last test (excluding inactive surfels moved by loop closures, which we only consider once they are active again to improve performance).
We delete detected invalid triangles, and also all other triangles connected to surfels within the neighbor search radii of their corner surfels (\cf Fig.~\ref{fig:remeshing_sketch}(b)). All affected surfels are scheduled for meshing in the following meshing iteration, \ie, the meshing algorithm in Sec.~\ref{sec:meshing} also fills the holes created by this step. This is possible since the remaining mesh can be used as initial state for meshing, and new triangles are added at mesh boundaries such as holes.

\section{Implementation Details}
\label{sec:implementation_details}
We implemented surfel reconstruction and denoising on a GPU using CUDA 8.0. 
Meshing and remeshing run in parallel on the CPU.
When a meshing iteration is expected to finish (based on the previous iteration's duration), we transfer the current surfel cloud from the GPU to the CPU. Camera poses and loop closures are computed via a state-of-the-art SLAM system~\cite{whelan2015elasticfusion}. 

\PAR{Depth image preprocessing.}
We run a number of preprocessing steps on the depth images before using them in our method.
These steps purely aim to improve the quality of the depth images.
They thus depend on the specific camera used (we mainly used the Kinect v1) and would not be required for very high-quality depth cameras.

We drop pixels with a depth larger than $3m$ since they tend to be very noisy~\cite{whelan2015elasticfusion}.
We also run a bilateral filter with depth-dependent $\sigma_{z}$ parameter ($\sigma_{xy} = 3 px, \sigma_{z} = 0.05 \cdot z$) on the images to mitigate quantization. Further, we filter outliers by projecting each pixel into the 4 previous and 4 following frames,  only keeping pixels that project to valid depth measurements within $2\%$ of the projected depth in all these frames. We remove all depth pixels within $2px$ of missing depth measurements to reduce the impact of foreground fattening.
As a final step, we compute normals via finite differences. Pixels whose normal differs by more than 85$^\circ$ from the direction to the camera are dropped.

In our evaluations, we only apply these preprocessing steps to our own method, since experiments showed that they generally do not improve the results of other methods.

\section{Evaluation}
We tested our method on a PC with an Intel Core i7 6700K and an MSI Geforce GTX 1080 Gaming X 8G.
We mainly evaluate our approach on the TUM RGB-D benchmark~\cite{sturm12iros} and the ICL-NUIM dataset~\cite{handa2014benchmark}, but also use the pre-registered version of the CoRBS dataset~\cite{wasenmueller2016corbs}.
Note that due to the nature of the topic, our approach is best seen in motion in the supplementary video (available at \url{https://youtu.be/CzMtNxuQ0OY}).

To refer back to the concepts of \emph{flexibility}, \emph{adaptiveness}, and \emph{efficiency} presented in the introduction, we highlight here how we evaluate for those:
We demonstrate \emph{flexibility} by showing how our approach handles loop closures in the supplementary video, and by showing results on sequences with loop closures in Tab.~\ref{tab:icl_nuim_accuracy} and Fig.~\ref{fig:qualitative_reconstructions}.
We demonstrate \emph{adaptiveness} among the qualitative results in Fig.~\ref{fig:triangulation} and also show how our approach handles input of different resolution in Fig.~\ref{fig:downsampling_texture_detail} and Fig.~\ref{fig:downsampling_geometry_detail}.
\emph{Efficiency} is evaluated in detail for different input resolutions in Fig.~\ref{fig:performance}.

\begin{table*}[t]
\caption{Mesh quality of our method (with and without regularization (reg), blending (bld), and incremental remeshing (remesh)) and of volume-based methods.
We evaluate the amount of unreferenced vertices (free), the amount of vertices on a mesh boundary (bdry), the average minimum triangle angle (angle), the amount of vertices having a locally manifold (boundary or non-boundary) triangle neighborhood with consistent orientation (manif), the amount of triangles having a self-intersection with the mesh (intsc), and the mean curvature (crv) in $\frac{0.01}{m}$.
}
\label{tab:mesh_quality}
\centering
\scriptsize
\begin{tabular}{ccc|c@{\hspace{0.7em}}c@{\hspace{0.7em}}c@{\hspace{0.7em}}c@{\hspace{0.7em}}c@{\hspace{0.7em}}c|c@{\hspace{0.7em}}c@{\hspace{0.7em}}c@{\hspace{0.7em}}c@{\hspace{0.7em}}c@{\hspace{0.7em}}c|c@{\hspace{0.7em}}c@{\hspace{0.7em}}c@{\hspace{0.7em}}c@{\hspace{0.7em}}c@{\hspace{0.7em}}c}
&& & \multicolumn{6}{c|}{fr1/desk~\cite{sturm12iros}} & \multicolumn{6}{c|}{fr1/xyz~\cite{sturm12iros}} & \multicolumn{6}{c}{fr3/office~\cite{sturm12iros}}\\
reg & bld & remesh & free & bdry & angle & manif & intsc & crv & free & bdry & angle & manif & intsc & crv & free & bdry & angle & manif & intsc & crv\\
\hline
O & O & O & 0.8\% & 4.8\% & 31.4$^\circ$\hspace{-0.2em} & 98.6\% & 0.4\% & 2.0 & 0.4\% & 3.6\% & 32.2$^\circ$\hspace{-0.2em} & 99.3\% & 0.2\% & 1.3 & 0.5\% & 2.8\% & \textbf{32.9$^\circ$\hspace{-0.2em}} & 99.2\% & 0.3\% & 1.4\\
O & O & X & 0.8\% & 4.8\% & \textbf{31.5$^\circ$\hspace{-0.2em}} & 98.4\% & 0.6\% & 2.0 & 0.5\% & 3.5\% & \textbf{32.5$^\circ$\hspace{-0.2em}} & 99.1\% & 0.2\% & 1.3 & 0.5\% & 2.7\% & 32.7$^\circ$\hspace{-0.2em} & 99.0\% & 0.5\% & 1.4\\
O & X & O & 0.3\% & 2.8\% & 30.9$^\circ$\hspace{-0.2em} & 99.4\% & 0.4\% & 1.1 & \textbf{0.1\%} & 2.0\% & 32.0$^\circ$\hspace{-0.2em} & 99.8\% & \textbf{0.1\%} & 0.6 & \textbf{0.1\%} & 1.6\% & 32.5$^\circ$\hspace{-0.2em} & 99.7\% & 0.2\% & 0.7\\
O & X & X & 0.3\% & 2.8\% & 31.3$^\circ$\hspace{-0.2em} & 99.2\% & 0.5\% & 1.1 & \textbf{0.1\%} & 1.9\% & \textbf{32.5$^\circ$\hspace{-0.2em}} & 99.8\% & \textbf{0.1\%} & 0.6 & 0.2\% & 1.5\% & 32.6$^\circ$\hspace{-0.2em} & 99.6\% & 0.2\% & 0.7\\
X & O & O & 0.7\% & 3.6\% & 29.1$^\circ$\hspace{-0.2em} & 99.0\% & 0.3\% & 1.0 & 0.2\% & 2.3\% & 31.0$^\circ$\hspace{-0.2em} & 99.7\% & \textbf{0.1\%} & 0.6 & 0.3\% & 2.0\% & 31.1$^\circ$\hspace{-0.2em} & 99.6\% & 0.2\% & 0.7\\
X & O & X & 0.7\% & 3.4\% & 29.5$^\circ$\hspace{-0.2em} & 98.9\% & 0.4\% & 1.0 & 0.2\% & 2.1\% & 31.7$^\circ$\hspace{-0.2em} & 99.6\% & 0.2\% & 0.6 & 0.3\% & 1.7\% & 31.3$^\circ$\hspace{-0.2em} & 99.5\% & 0.2\% & 0.7\\
X & X & O & \textbf{0.2\%} & 2.4\% & 29.5$^\circ$\hspace{-0.2em} & \textbf{99.6\%} & \textbf{0.2\%} & \textbf{0.6} & \textbf{0.1\%} & 1.9\% & 31.4$^\circ$\hspace{-0.2em} & \textbf{99.9\%} & \textbf{0.1\%} & \textbf{0.3} & \textbf{0.1\%} & 1.5\% & 31.7$^\circ$\hspace{-0.2em} & \textbf{99.8\%} & \textbf{0.1\%} & \textbf{0.5}\\
\textbf{X} & \textbf{X} & \textbf{X} & \textbf{0.2\%} & 2.3\% & 30.0$^\circ$\hspace{-0.2em} & 99.5\% & 0.3\% & \textbf{0.6} & \textbf{0.1\%} & 1.7\% & 32.0$^\circ$\hspace{-0.2em} & 99.8\% & \textbf{0.1\%} & \textbf{0.3} & \textbf{0.1\%} & 1.3\% & 32.0$^\circ$\hspace{-0.2em} & 99.7\% & 0.2\% & \textbf{0.5}\\
\hline
\multicolumn{3}{l|}{InfiniTAM \cite{infinitam2015ismar}} & \textbf{0.0\%} & 9.7\% & 32.5$^\circ$\hspace{-0.2em} & \textbf{100\%} & \textbf{0.0\%} & 7.4 & \textbf{0.0\%} & 11.1\% & 33.7$^\circ$\hspace{-0.2em} & \textbf{100\%} & \textbf{0.0\%} & 3.9 & \textbf{0.0\%} & 5.5\% & \textbf{35.3$^\circ$\hspace{-0.2em}} & \textbf{100\%} & \textbf{0.0\%} & 2.6\\
\multicolumn{3}{l|}{FastFusion \cite{steinbruecker2014volumetric}} & \textbf{0.0\%} & 17.7\% & \textbf{33.8$^\circ$\hspace{-0.2em}} & 98.4\% & 1.0\% & 8.4 & \textbf{0.0\%} & 15.6\% & \textbf{34.1$^\circ$\hspace{-0.2em}} & 99.1\% & 0.6\% & 7.2 & \textbf{0.0\%} & 18.3\% & 34.8$^\circ$\hspace{-0.2em} & 97.9\% & 1.7\% & 6.6\\
\end{tabular}
\end{table*}

\begin{figure}[!t]
\centering
\renewcommand{\tabcolsep}{1px}
\renewcommand{\arraystretch}{0}
\begin{tabular}{cc}
\begin{minipage}{0.499\linewidth}\includegraphics[trim={0 37px 0 35px},clip,width=1\linewidth]{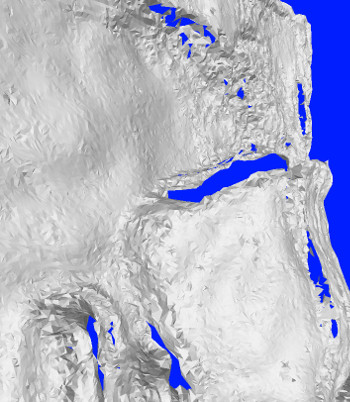}\end{minipage}&\begin{minipage}{0.499\linewidth}\includegraphics[trim={0 37px 0 35px},clip,width=1\linewidth]{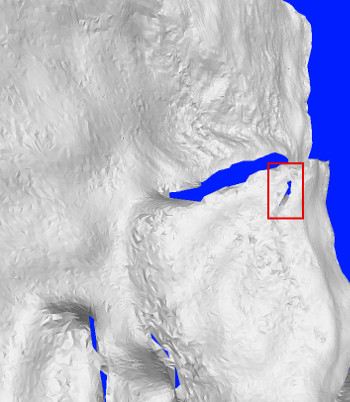}\end{minipage}
\end{tabular}
~\vspace{0.28em}~\\
\begin{tabular}{cc}
\begin{minipage}{0.499\linewidth}\includegraphics[trim={0 37px 0 35px},clip,width=1\linewidth]{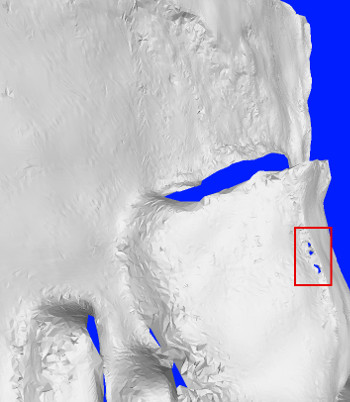}\end{minipage}&\begin{minipage}{0.499\linewidth}\includegraphics[trim={0 37px 0 35px},clip,width=1\linewidth]{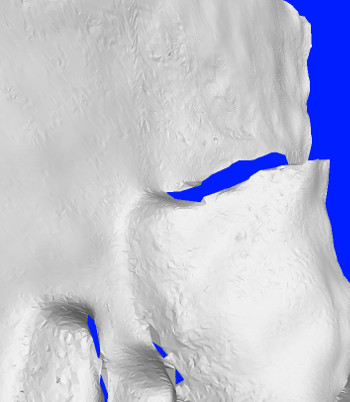}\end{minipage}\end{tabular}
\caption{Comparison between no smoothing (top left), observation boundary blending only (top right), regularization only (bottom left), and both (bottom right).
Both types of smoothing contribute to a hole-free triangulation and improve the surface quality.}
\label{fig:denoising_evaluation}
\end{figure}

\begin{figure*}[!t]
\centering
\renewcommand{\tabcolsep}{0px}
\renewcommand{\arraystretch}{0}
\newcommand{\imagewidth}{0.172\linewidth}
\begin{tabular}{cc@{\hspace{1em}}|@{\hspace{1em}}ccc}
\includegraphics[trim={0 0 0 0},clip,width=\imagewidth]{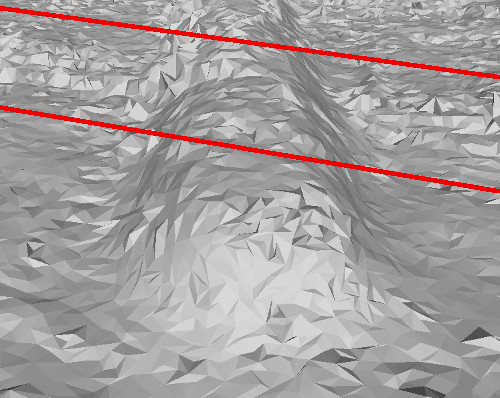}&\includegraphics[trim={0 0 0 0},clip,width=\imagewidth]{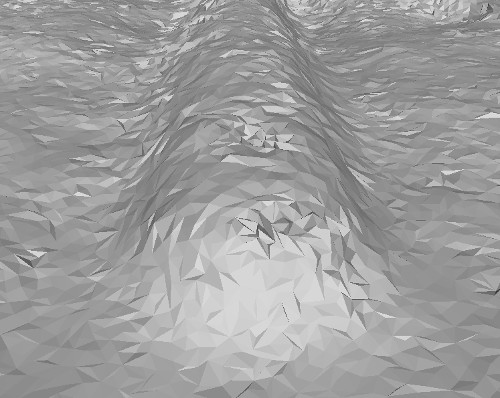}&\includegraphics[trim={0 0 0 0},clip,width=\imagewidth]{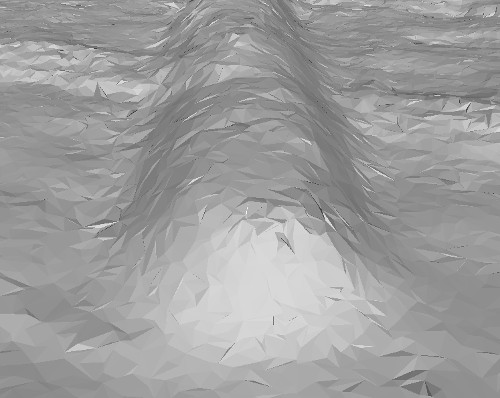}&\includegraphics[trim={0 0 0 0},clip,width=\imagewidth]{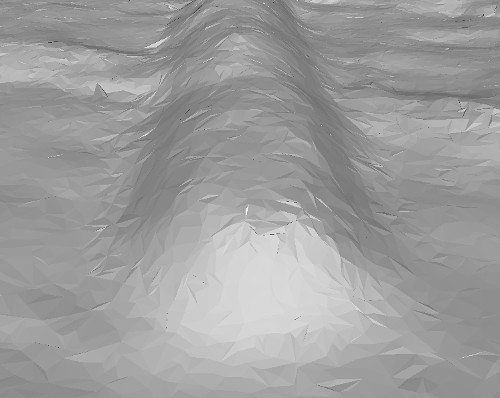}&\includegraphics[trim={0 0 0 0},clip,width=\imagewidth]{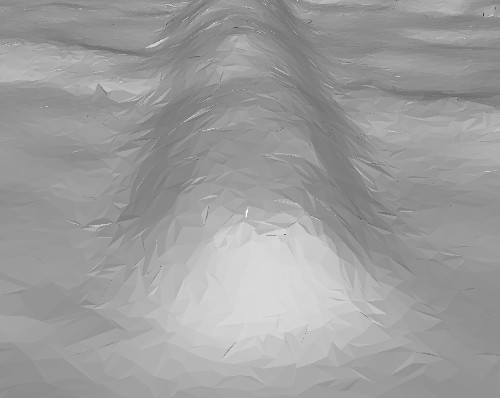}\\%
\includegraphics[trim={0 0 0 0},clip,width=\imagewidth]{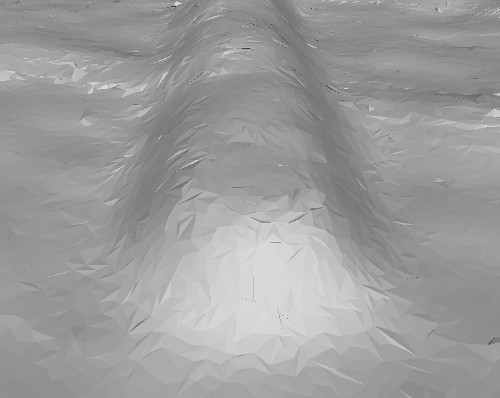}&\includegraphics[trim={0 0 0 0},clip,width=\imagewidth]{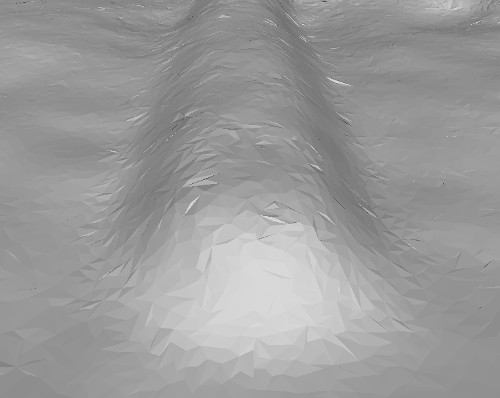}&\includegraphics[trim={0 0 0 0},clip,width=\imagewidth]{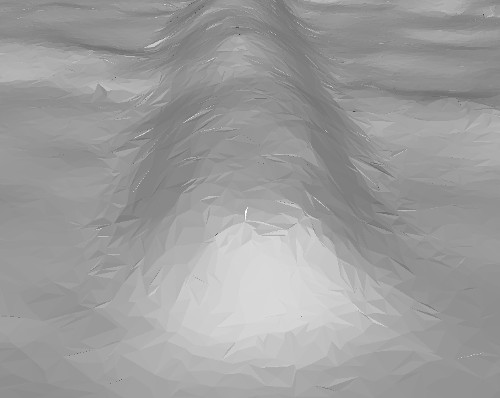}&\includegraphics[trim={0 0 0 0},clip,width=\imagewidth]{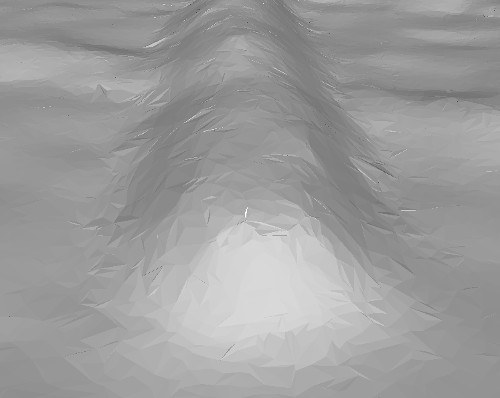}&\includegraphics[trim={0 0 0 0},clip,width=\imagewidth]{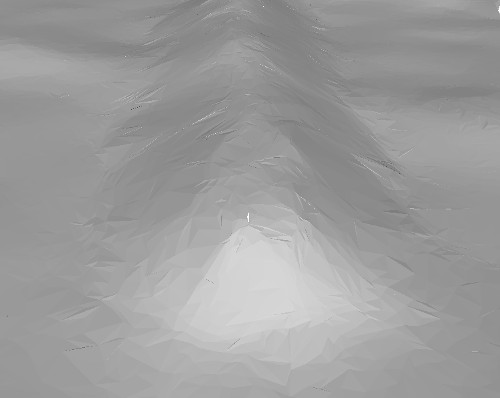}\end{tabular}
\caption{Comparison between our denoising (with regularization, shown in the fully converged state) and standard Laplacian Smoothing.
\textbf{Left block:} no smoothing (top left), observation boundary blending only (top right), regularization only (bottom left), and both (bottom right).
\textbf{Right block:} Top row: one, two, three iterations of Laplacian smoothing only. Bottom row: four, five, twenty iterations of Laplacian smoothing only.
Out of all smoothing methods, only boundary blending is able to remove the horizontal scanning artefact caused by small pose errors (marked in red in the top left image).
Laplacian smoothing shrinks the object and continues to smooth it with every iteration, while ours converges to a good solution due to our data term.}
\label{fig:denoising_vs_laplacian_smoothing}
\end{figure*}

\subsection{Ablation Study}
\label{sec:ablation_study}

\PAR{The impact of denoising.}
In a first experiment, we show that the two denoising / smoothing steps (regularization and boundary blending) introduced in Sec.~\ref{sec:denoising} are necessary when meshing surfel point clouds. 

Fig.~\ref{fig:denoising_evaluation} shows that both proposed smoothing steps contribute to a hole-free mesh.
In this example, both steps are required to get a closed surface. 
We also compare our smoothing against standard Laplacian smoothing.
Laplacian smoothing is defined by the update equation $\mathbf{p}_s := \mathbf{p}_s + \lambda /|N_s| \sum_{n \in N_s} (\mathbf{p}_n - \mathbf{p}_s)$.
Notice that Laplacian smoothing does not contain a data term and is thus prone to shrinking surfaces towards the mean of all data points. 
We use the same vertex neighbor set for $N_s$ which we use for our regularization approach, and choose $\lambda = 1$.
Fig.~\ref{fig:denoising_vs_laplacian_smoothing} shows that in contrast to standard Laplacian smoothing, our approach avoids object shrinkage and does not over-smooth when many iterations are applied.
Furthermore, only our blending step is able to remove a scanning boundary artefact.

As a quantitative experiment, Tab.~\ref{tab:mesh_quality} evaluates a number of mesh quality metrics with and without our denoising steps on the reconstructions of three datasets.
Both denoising steps consistently yield smoother surfaces on these datasets when enabled.
In addition, they reduce the number of free and boundary surfels, indicating less noise and holes.
Furthermore, meshes are manifold in more places and there are fewer self-intersections. 
This clearly shows that our proposed denoising stages lead to better surface reconstructions.

\begin{figure*}[t]
\begin{center}
\begin{overpic}[trim={0 11pt 0 6pt},clip,width=0.99\linewidth]{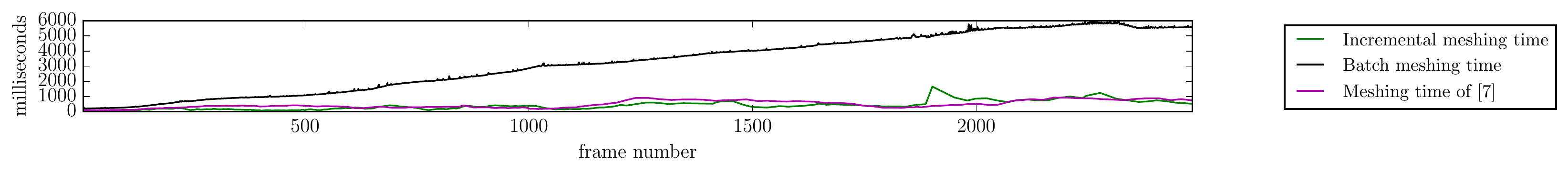}
 \put (94.2,3.2) {\textcolor{white}{\rule{0.3cm}{0.3cm}}}
 \put (94.0,3.7) {\scalebox{1.2}{\tiny\cite{steinbruecker2014volumetric}}}
\end{overpic}
\end{center}
\caption{Performance of our incremental meshing (sum of synchronization, remeshing and meshing times per frame) on the dataset from Fig.~\ref{fig:teaser} compared to: 1) Batch-meshing the surfel reconstruction at each frame from scratch, and 2) \cite{steinbruecker2014volumetric}'s meshing performance.}
\label{fig:performance_comparison}
\end{figure*}

\PAR{The impact of remeshing.}
Remeshing is required for updating existing surfaces.
To be more effective than meshing from scratch, it must be faster while providing comparable results.
We show the latter by comparing the mesh quality of the reconstructions generated by our proposed approach with remeshing (lines with ``remesh'' in Tab.~\ref{tab:mesh_quality}) with an approach that meshes the final surfel clouds from scratch (lines without ``remesh'' in Tab.~\ref{tab:mesh_quality}).
Most metrics are similar, with the meshing from scratch yielding results that are manifold in slightly more places and that have slightly less self-intersections.
The latter is to be expected, since we do not test for this in the remeshing algorithm.
Notice that in our implementation of \cite{gopi2000fast,marton2009fast}, we favor completeness over manifoldness. 
Still, the overall differences between remeshing and meshing from scratch are very small.
Meshing the surfel cloud from scratch is much slower than our partial remeshing, as shown in Fig.~\ref{fig:performance_comparison}.
For example, meshing the final surfel cloud in this example from scratch takes about $5.6$ seconds. 
Fig.~\ref{fig:performance_comparison} also shows that the remeshing time of the volumetric FastFusion approach from \cite{steinbruecker2014volumetric} (purple line) is similar to ours.

Overall, our experiments clearly demonstrate that our online remeshing approach achieves comparable results as performing meshing from scratch at faster run-times and better scalability, since the performance mostly depends on the region affected by remeshing instead of the size of the whole reconstruction.

\begin{figure*}[t!]
\begin{center}
\begin{minipage}{0.375\linewidth}
\includegraphics[trim={0 2.2em 0 0.7em},clip,width=1.0\linewidth]{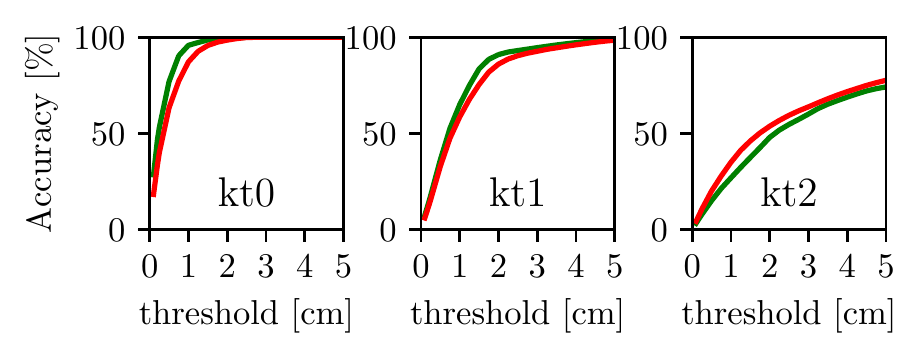}\\%
\includegraphics[trim={0 0 0 0.7em},clip,width=1.0\linewidth]{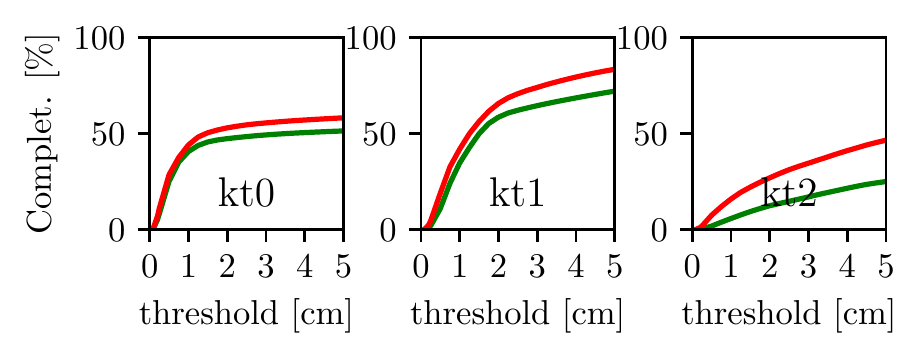}\end{minipage}\begin{minipage}{0.475\linewidth}
\includegraphics[trim={0 2.2em 0 0.7em},clip,width=1.0\linewidth]{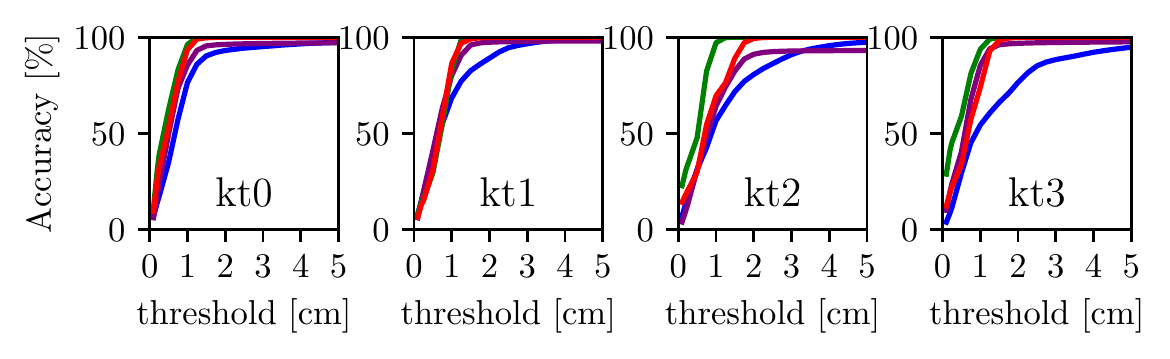}\\
\includegraphics[trim={0 0 0 0.7em},clip,width=1.0\linewidth]{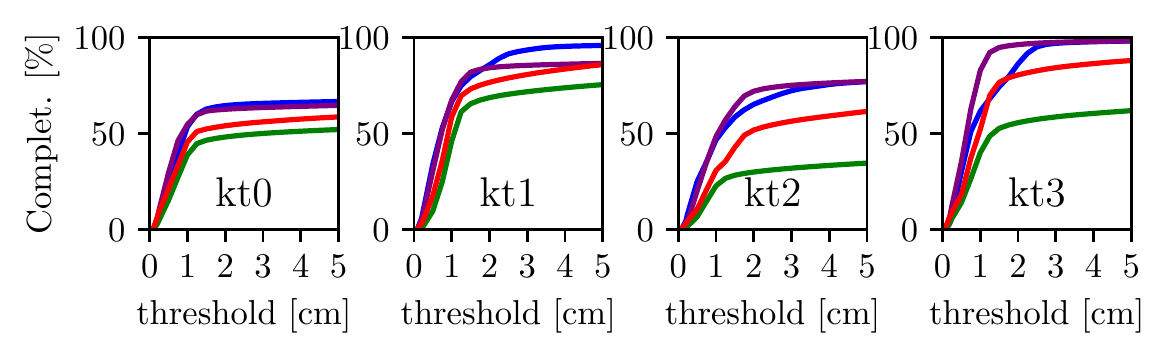}
\end{minipage}\begin{minipage}{0.15\linewidth}
\begin{overpic}[trim={5.4em 4.4em 5.4em 4.4em},clip,width=1.0\linewidth]{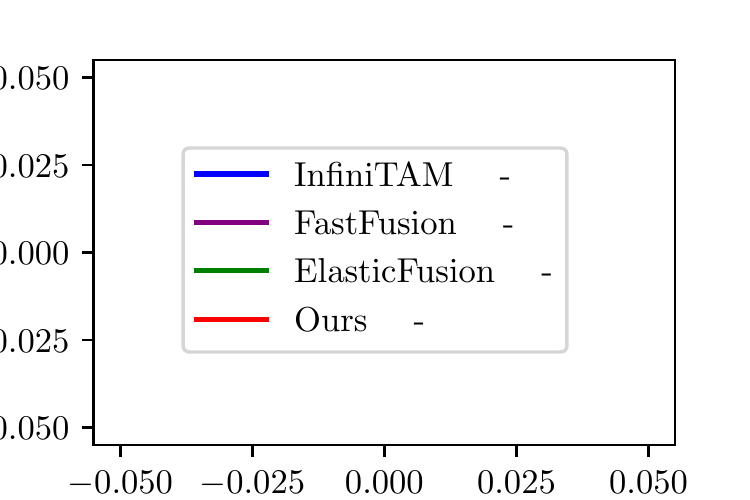}
 \put (55,5) {\textcolor{white}{\rule{0.3cm}{0.3cm}}}
 \put (85,18) {\textcolor{white}{\rule{0.3cm}{0.3cm}}}
 \put (80.5,18.3) {\scalebox{1.32}{\tiny\cite{whelan2015elasticfusion}}}
 \put (80,29) {\textcolor{white}{\rule{0.3cm}{0.3cm}}}
 \put (71.5,30.6) {\scalebox{1.32}{\tiny\cite{steinbruecker2014volumetric}}}
 \put (78,40) {\textcolor{white}{\rule{0.3cm}{0.3cm}}}
 \put (71,43) {\scalebox{1.32}{\tiny\cite{infinitam2015ismar}}}
\end{overpic}
\end{minipage}\end{center}
\caption{Accuracy and completeness from Tab.~\ref{tab:icl_nuim_accuracy}, plotted for varying evaluation threshold.
Left: trajectory with loop closures, right: ground truth trajectory.
}
\label{fig:icl_nuim_plots}
\end{figure*}

\subsection{Comparison with the State-of-the-Art}
\label{sec:evaluation:comparison}

\PAR{Mesh quality.} 
Tab.~\ref{tab:mesh_quality} also compares the mesh quality of our approach with two volume-based approaches, InfiniTAM \cite{infinitam2015ismar} and FastFusion \cite{steinbruecker2014volumetric}.
We use source code provided by the authors for comparison.
Both approaches use the Marching Cubes algorithm \cite{lorensen1987marching} for obtaining a triangle mesh from their volumetric representations.
We merge vertices having the same position before evaluation.
By design, Marching Cubes produces clean meshes wrt.~unreferenced vertices, manifoldness and self-intersections (the meshing of \cite{steinbruecker2014volumetric} seems to be affected by bugs though).
However, judging by the minimum triangle angles, our method creates triangles of comparable quality. 
Our approach creates less boundary vertices and our meshes exhibit a significantly lower curvature, indicating smoother reconstructions. 
At the same time, our approach is flexible enough to adapt to loop closure events (see also the supplementary video), which cannot be handled by \cite{infinitam2015ismar,steinbruecker2014volumetric}.

\begin{table}[t!]
\centering
\caption{Accuracy [\%], completeness [\%], and mean curvature [$\frac{0.01}{m}$] results for ICL-NUIM \cite{handa2014benchmark} sequences.
Evaluation threshold $1cm$.
Our method generally yields higher completeness than ElasticFusion \cite{whelan2015elasticfusion} and higher accuracy than InfiniTAM \cite{infinitam2015ismar} and \cite{steinbruecker2014volumetric}.
Since \cite{infinitam2015ismar} and \cite{steinbruecker2014volumetric} cannot handle loop closures, there are no results for these cases.
For results tagged ``smoothed'', we applied the same bilateral filter to the input as we do within the preprocessing for our method (\cf Sec.~\ref{sec:implementation_details}).
}
\label{tab:icl_nuim_accuracy}
\scriptsize
\begin{tabular}{@{\hspace{0.72em}}c@{\hspace{0.72em}}|@{\hspace{0.72em}}l@{\hspace{0.72em}}|@{\hspace{0.72em}}c@{\hspace{0.72em}}c@{\hspace{0.72em}}c@{\hspace{0.72em}}|@{\hspace{0.72em}}c@{\hspace{0.72em}}c@{\hspace{0.72em}}c@{\hspace{0.72em}}c@{\hspace{0.72em}}}
&              & \multicolumn{3}{@{\hspace{0.72em}}c@{\hspace{0.72em}}|@{\hspace{0.72em}}}{with}                 & \multicolumn{4}{@{\hspace{0.72em}}c@{\hspace{0.72em}}}{ground truth} \\
&              & \multicolumn{3}{@{\hspace{0.72em}}c@{\hspace{0.72em}}|@{\hspace{0.72em}}}{loop closures}                 & \multicolumn{4}{@{\hspace{0.72em}}c@{\hspace{0.72em}}}{trajectory} \\
& {\bf Method} & {\bf kt0} & {\bf kt1} & {\bf kt2}                                          & {\bf kt0} & {\bf kt1} & {\bf kt2} & {\bf kt3} \\
\hline
\parbox[t]{2mm}{\multirow{7}{*}{\rotatebox[origin=c]{90}{Accuracy}}} & InfiniTAM \cite{infinitam2015ismar}    & - & - & -                                       & 76.4 & 68.4 & 56.6 & 54.4 \\
 & InfiniTAM \cite{infinitam2015ismar} - smoothed    & - & - & -                            & 78.3 & 68.3 & 58.1 & 58.3\\
 & FastFusion \cite{steinbruecker2014volumetric}     & - & - & -                            & 85.5 &  80.1 &  64.2 &  85.3 \\
 & FastFusion \cite{steinbruecker2014volumetric} - smoothed     & - & - & -                 & 75.9 & 78.8 & 52.6 & 72.5\\
 & ElasticFusion \cite{whelan2015elasticfusion} & 95.8 & \textbf{64.9} &  26.9     & \textbf{96.2} &  83.1 & \textbf{97.1} & \textbf{93.7} \\
 & ElasticFusion \cite{whelan2015elasticfusion} - smoothed & \textbf{96.8} & 64.4 & 26.8             & 95.7 & 82.1 & 96.6 & 92.2\\
 & \textbf{SurfelMeshing (Ours)} &  87.2 &  58.5 & \textbf{35.0}                            & 93.5 & \textbf{86.4} &  69.5 &  74.0 \\
\hline
\parbox[t]{2mm}{\multirow{7}{*}{\rotatebox[origin=c]{90}{Completeness}}} & InfiniTAM \cite{infinitam2015ismar}    & - & - & -                                       & 53.5 & 66.5 & 46.7 & 61.6 \\
 & InfiniTAM \cite{infinitam2015ismar} - smoothed    & - & - & -                            & 51.6 & 62.8 & 42.6 & 59.9\\
 & FastFusion \cite{steinbruecker2014volumetric}     & - & - & -                            & \textbf{54.7} & \textbf{67.3} & \textbf{48.4} & \textbf{82.8} \\
 & FastFusion \cite{steinbruecker2014volumetric} - smoothed     & - & - & -                 & 45.6 & 63.0 & 38.8 & 66.5\\
 & ElasticFusion \cite{whelan2015elasticfusion} &  40.6 &  34.6 &   5.8                     & 38.8 & 46.0 & 22.8 & 40.0 \\
 & ElasticFusion \cite{whelan2015elasticfusion} - smoothed &  41.8 & 34.5 & 5.8             & 38.9 & 45.3 & 23.1 & 40.4\\
 & \textbf{SurfelMeshing (Ours)} & \textbf{44.0} & \textbf{41.8} & \textbf{15.9}            & 45.6 &  58.6 &  30.8 &  52.1 \\
\hline
\parbox[t]{2mm}{\multirow{7}{*}{\rotatebox[origin=c]{90}{Curvature}}} & InfiniTAM \cite{infinitam2015ismar}    & - & - & -                                       & 2.48 & 2.23 & 4.71 & 3.69 \\
 & InfiniTAM \cite{infinitam2015ismar} - smoothed    & - & - & -                            & 1.46 & 0.93 & 1.71 & 1.34\\
 & FastFusion \cite{steinbruecker2014volumetric}     & - & - & -                            & 0.99 & 1.47 & 1.68 & 1.33\\
 & FastFusion \cite{steinbruecker2014volumetric} - smoothed     & - & - & -                 & 0.63 & 0.87 & 1.20 & 0.92\\
 & ElasticFusion \cite{whelan2015elasticfusion} & 0.86 & 0.39 & 0.39                        & 0.24 & 0.43 & 0.34 & 0.47 \\
 & ElasticFusion \cite{whelan2015elasticfusion} - smoothed & 0.37 & 0.30 & 0.35             & 0.18 & 0.35 & 0.28 & 0.41\\
 & \textbf{SurfelMeshing (Ours)} & \textbf{0.22} & \textbf{0.15} & \textbf{0.30}            & \textbf{0.15} & \textbf{0.17} & \textbf{0.18} & \textbf{0.32} \\
\end{tabular}
\end{table}

\PAR{Accuracy and completeness.}
We compare our reconstructions to other state-of-the-art methods in a quantitative evaluation. 
For this experiment, we use the living room sequences of the synthetic ICL-NUIM dataset \cite{handa2014benchmark}, which provides depth maps with simulated noise. 

\begin{figure}[!t]
\centering
\renewcommand{\tabcolsep}{1px}
\renewcommand{\arraystretch}{0}
\begin{tabular}{cc}
\includegraphics[trim={60px 4px 60px 4px},clip,width=0.49\linewidth]{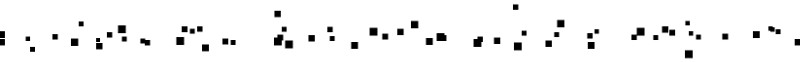}&\includegraphics[trim={60px 0 60px 0},clip,width=0.49\linewidth]{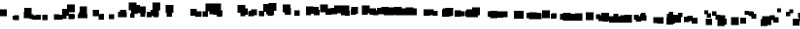}\\
\includegraphics[trim={60px 0px 60px 0px},clip,width=0.49\linewidth]{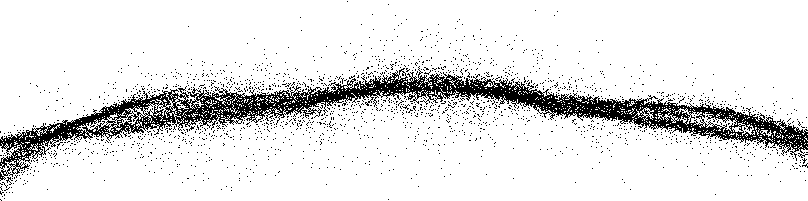}&\includegraphics[trim={60px 0px 60px 0px},clip,width=0.49\linewidth]{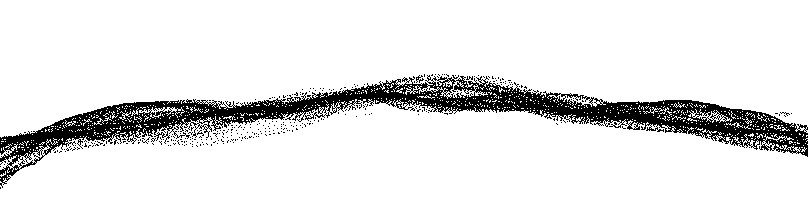}\end{tabular}
\caption{Cut through surfaces from ElasticFusion (top left) and \cite{lefloch2017comprehensive} (bottom left) compared to corresponding surfels in our approach (right). 
Our method generates smoother surfaces due to our denoising steps (Sec.~\ref{sec:denoising}).
}
\label{fig:denoising_cut}
\end{figure}

We first compare against ElasticFusion~\cite{whelan2015elasticfusion} to show that we improve upon it. The kt0 and kt2 sequences trigger loop closures. We leave out kt3 since the open source code of \cite{whelan2015elasticfusion} failed to estimate a reasonable trajectory. We align the reconstructions to the ground truth model with point-to-plane ICP. 
We compute accuracy / completeness as the amount of reconstructed surfels / ground truth points which are closer to the ground truth / reconstructed surfels than a given evaluation threshold \cite{Knapitsch2017,schoeps2017cvpr}.
We also evaluate mean curvature as a measure of smoothness.
Results are given in Tab.~\ref{tab:icl_nuim_accuracy} (left) for an evaluation threshold of $1cm$, and are plotted for different thresholds in Fig.~\ref{fig:icl_nuim_plots}.
ElasticFusion uses more aggressive outlier filtering and thus gives partly more accurate but always less complete results than ours. The plots show that this does not depend on the choice of evaluation threshold.
Our results are always smoother than ElasticFusion's due to our denoising. This is also shown qualitatively in Fig.~\ref{fig:denoising_cut}.

Next, we compare to the voxel-based methods \cite{steinbruecker2014volumetric} and InfiniTAM \cite{infinitam2015ismar}.
Comparing against methods that handle loop closures, \eg, \cite{kahler2016real}, is difficult without including the respective SLAM systems in the comparison. 
For this experiment, we therefore use the ground truth trajectories and disable loop closure handling for all methods.
Results are given in Tab.~\ref{tab:icl_nuim_accuracy} (right) and plotted in Fig.~\ref{fig:icl_nuim_plots}.
While the voxel-based methods can rely on the TSDF for outlier filtering and obtain more complete results, ours achieves higher accuracy almost throughout.
The plots show that these results are again mostly constant for varying the evaluation threshold within a reasonable range.
Furthermore, ours has several advantages due to not using a volume (\cf Sec.~\ref{sec:evaluation:qualitative}).
 
In Tab.~\ref{tab:icl_nuim_accuracy}, we also include results for the baseline methods on input that was smoothed with the same bilateral filter that we use in the preprocessing for our method (\cf Sec.~\ref{sec:implementation_details}).
For InfiniTAM, we observe that the smoothing shifts the accuracy-completeness tradeoff slightly in favor of accuracy, and strongly reduces the mean curvature of the reconstruction.
For FastFusion, smoothing causes a strong reduction in completeness.
Inspecting the reconstructions showed that some parts are erroneously missing with this kind of input, which appears to be a bug in the software.
For ElasticFusion, the smoothing appears to have little effect on the accuracy and completeness results, while clearly reducing the mean curvature.
In sum, while the smoothing has a varying effect on the accuracy and completeness measures, it significantly reduces the mean curvature for all methods, however without reaching the values of our approach.

\subsection{Qualitative Results}
\label{sec:evaluation:qualitative}

\begin{figure}[tb!]
\begin{center}
\includegraphics[trim={0 0 0 0},clip,width=0.5\linewidth]{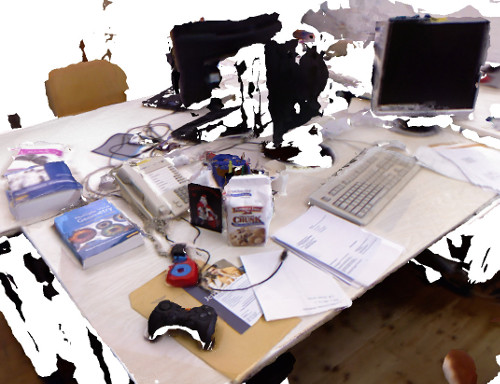}\includegraphics[trim={0 0 0 0},clip,width=0.5\linewidth]{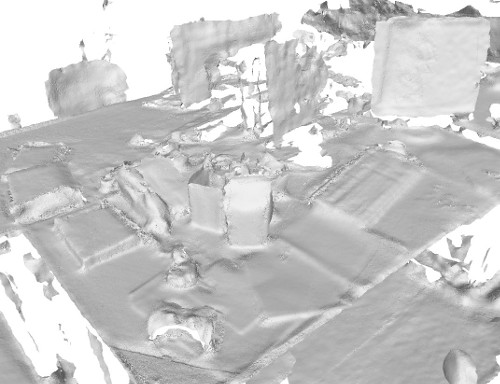}\\%
\includegraphics[trim={0 0 0 0},clip,width=0.5\linewidth]{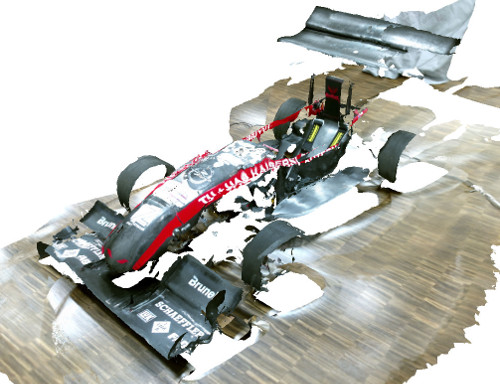}\includegraphics[trim={0 0 0 0},clip,width=0.5\linewidth]{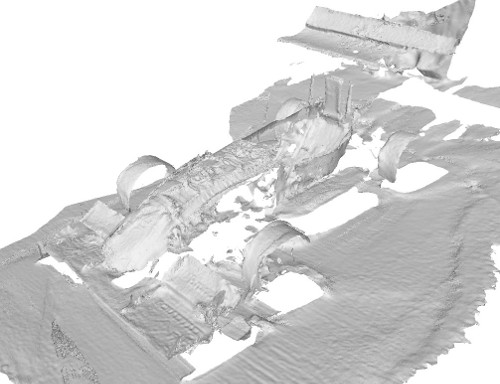}\\%
\includegraphics[trim={0 0 0 0},clip,width=0.5\linewidth]{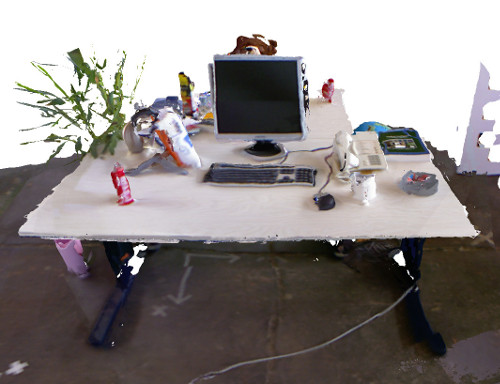}\includegraphics[trim={0 0 0 0},clip,width=0.5\linewidth]{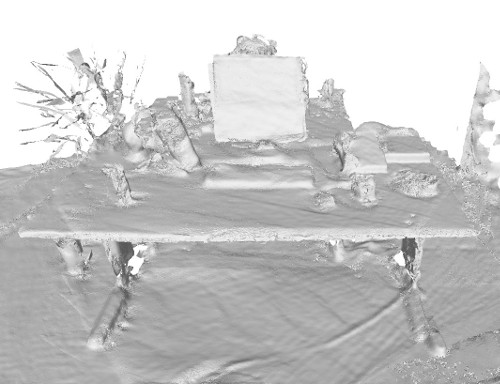}\\%
\includegraphics[trim={0 0 0 0},clip,width=0.25\linewidth]{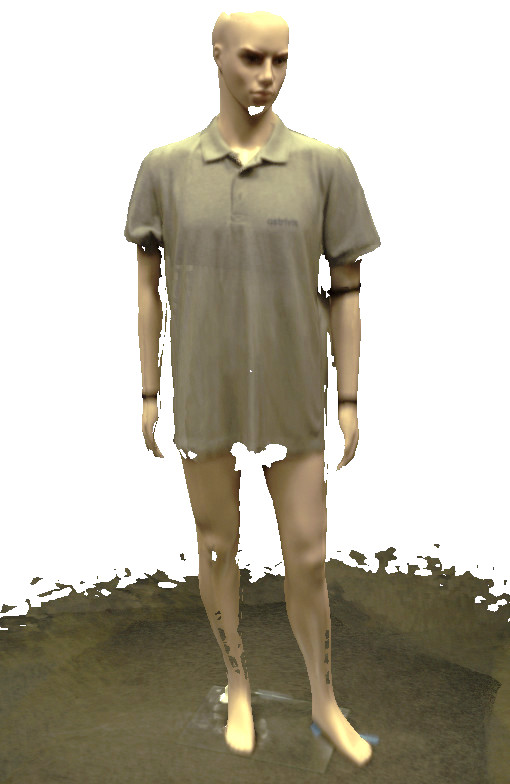}\includegraphics[trim={0 0 0 0},clip,width=0.25\linewidth]{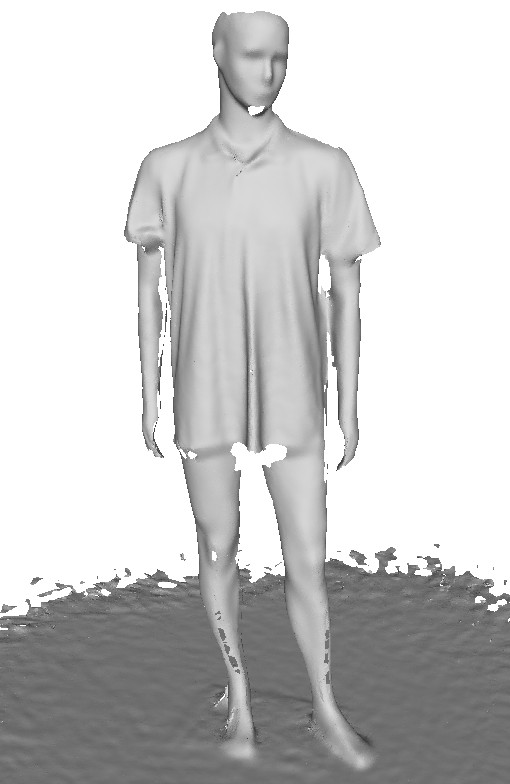}\includegraphics[trim={0 0 0 0},clip,width=0.25\linewidth]{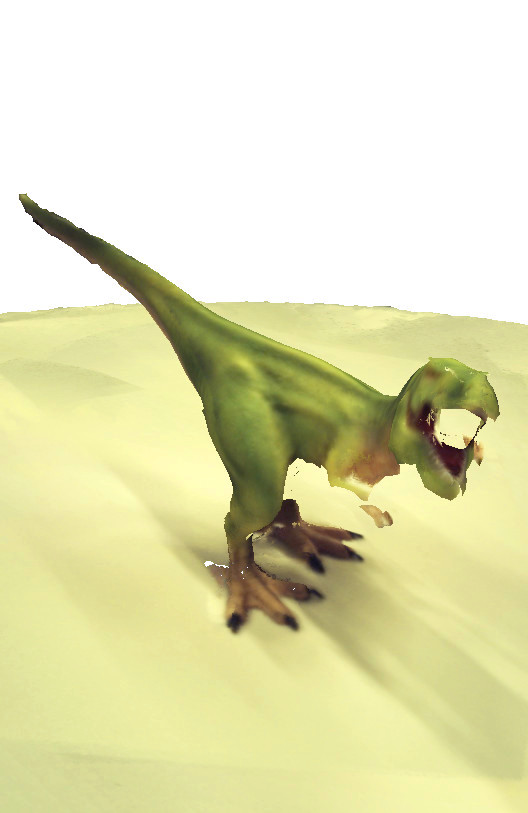}\includegraphics[trim={0 0 0 0},clip,width=0.25\linewidth]{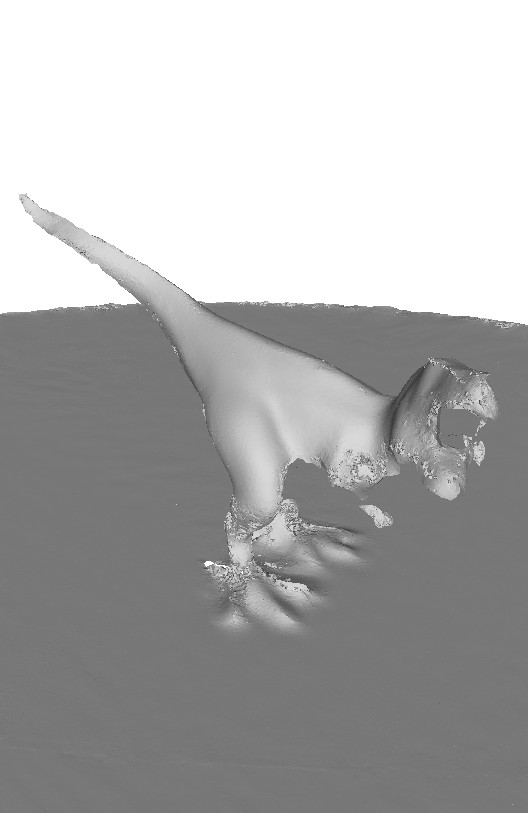}\\%
\end{center}
\caption[]{Qualitative results for our method (top to bottom; M = million):\\
fr1/desk \cite{sturm12iros}:~~~\hspace{0.15em}573 frames, 1.2M surfels, 2.4M triangles.\\
R3 \cite{wasenmueller2016corbs}:~~~~~~~~~~2151 frames, 1.9M surfels, 3.7M triangles.\\
fr2/desk \cite{sturm12iros}:~~2893 frames, 1.3M surfels, 2.6M triangles.\\
Last row, left to right, with camera poses provided by BAD SLAM \cite{schoeps2019badslam}:\\
eth3d/mannequin1 \cite{schoeps2019badslam}:~~~~643 frames, 1.3M surfels, 2.4M triangles.\\
eth3d/dino \cite{schoeps2019badslam}:~~~~~~~~~~~~~~~2074 frames, 0.6M surfels, 1.3M triangles.\\
}
\label{fig:qualitative_reconstructions}
\end{figure}

We perform qualitative experiments on real datasets, mainly using datasets from \cite{sturm12iros} captured with Kinect v1 cameras.
In addition to Fig.~\ref{fig:teaser}, Fig.~\ref{fig:qualitative_reconstructions} qualitatively shows reconstructions for different scenes, with statistics of the datasets and reconstructions.
All estimated camera trajectories for these datasets include loop closures.

\begin{figure}[t]
\renewcommand{\tabcolsep}{0px}
\renewcommand{\arraystretch}{0}
\begin{tabular}{ccc@{\hspace{0.2em}}|@{\hspace{0.2em}}c}
\includegraphics[trim={0 0px 0 0px},clip,width=0.244\linewidth]{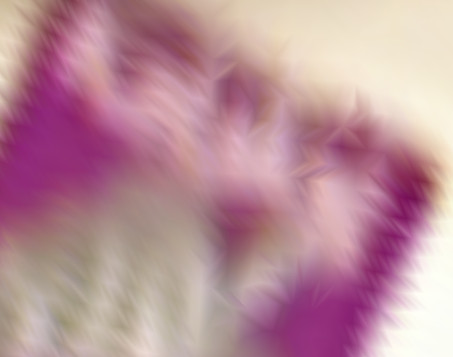}&\includegraphics[trim={0 0px 0 0px},clip,width=0.244\linewidth]{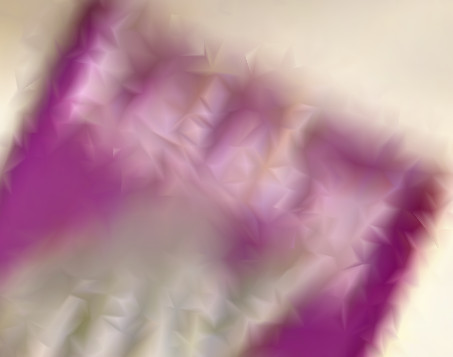}&\includegraphics[trim={0 0px 0 0px},clip,width=0.244\linewidth]{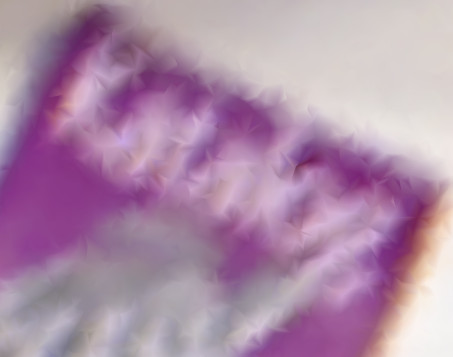}&\includegraphics[trim={0 0px 0 12px},clip,width=0.244\linewidth]{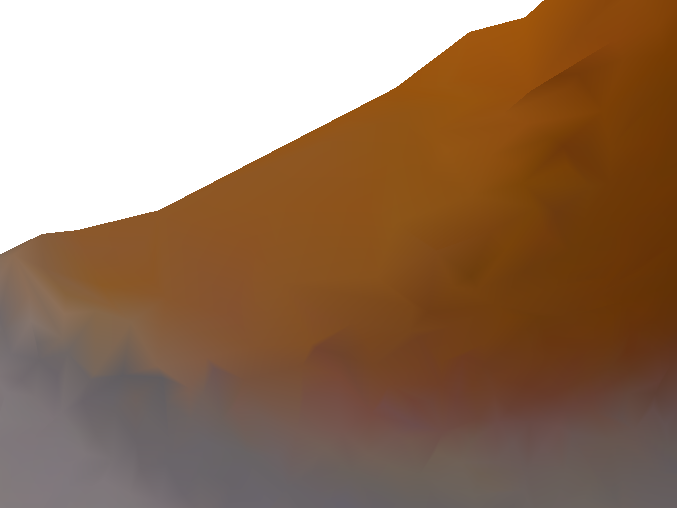}\\%
\includegraphics[trim={0 0px 0 0px},clip,width=0.244\linewidth]{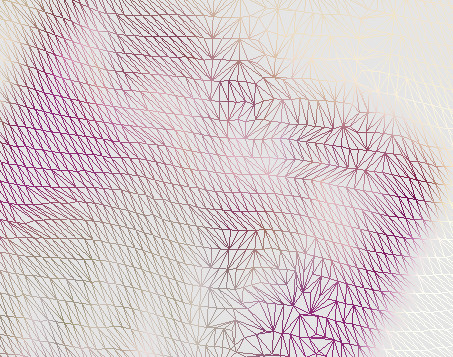}&\includegraphics[trim={0 0px 0 0px},clip,width=0.244\linewidth]{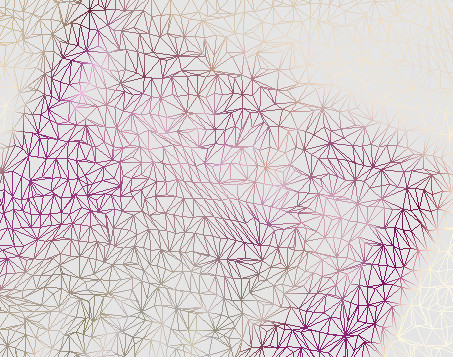}&\includegraphics[trim={0 0px 0 0px},clip,width=0.244\linewidth]{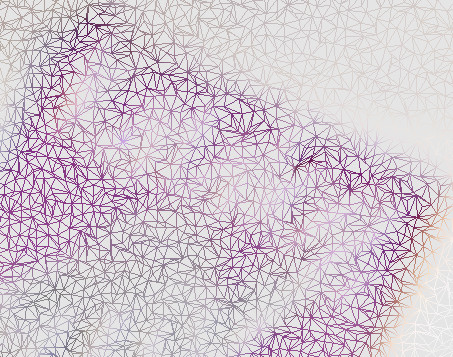}&\includegraphics[trim={0 0px 0 12px},clip,width=0.244\linewidth]{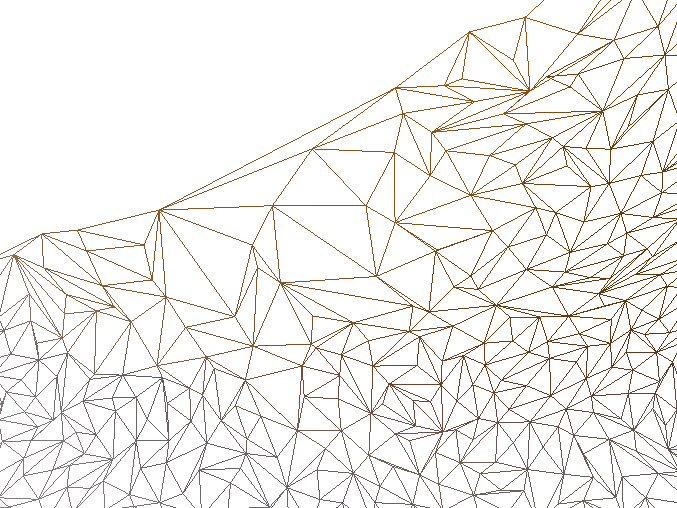}\end{tabular}
\caption{\textbf{Left:} As the camera moves closer (left to right), the mesh (bottom) is refined and thus the colors (top) become more detailed.
\textbf{Right:} The algorithm naturally supports triangulations with differing surfel resolution.}\label{fig:triangulation}
\end{figure}

\PAR{Adaptivity to varying scan resolution.}
Since surfels are created to match the input image resolution, our approach reconstructs the scene's colors at this resolution without texture mapping. This is illustrated by Fig.~\ref{fig:triangulation} (left), which shows the mesh being refined as the camera moves closer.
Voxel-based methods require a very high volume resolution to match this capability.
Thus, they typically reconstruct blurrier texture, as shown in Fig.~\ref{fig:representation_comparison}.
This figure also shows that our approach leads to improved sharpness compared to the surfel-based ElasticFusion method \cite{whelan2015elasticfusion}. 
This is due to denoising (\cf Fig.~\ref{fig:denoising_cut}) and meshing, which improves the appearance over individual splat rendering.
In our experience, transitions between areas of different resolution do not cause triangulation issues (\cf Fig.~\ref{fig:triangulation} (right)).

\begin{figure}[!tbp]
\begin{center}
\renewcommand{\tabcolsep}{0px}
\renewcommand{\arraystretch}{0}
\begin{tabular}{cccc}
\includegraphics[trim={0 0 0 0},clip,width=0.249\linewidth]{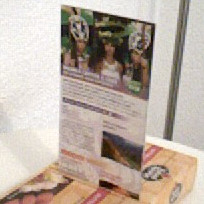}&\includegraphics[trim={0 0 0 0},clip,width=0.249\linewidth]{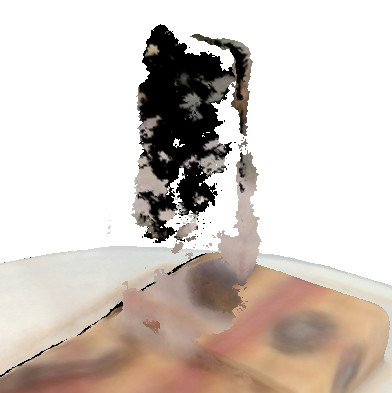}&\includegraphics[trim={0 0 0 0},clip,width=0.249\linewidth]{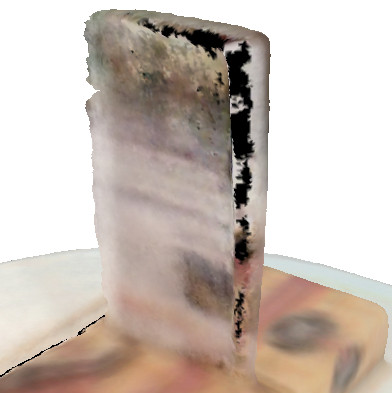}&\includegraphics[trim={0 0 0 0},clip,width=0.249\linewidth]{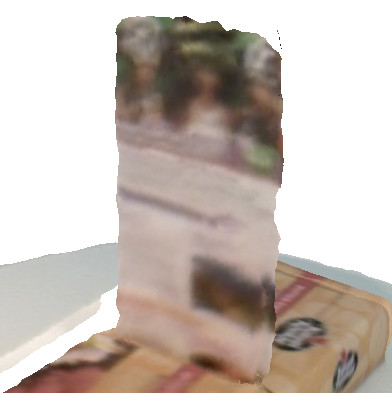}\\%
\includegraphics[trim={0 0 0 0},clip,width=0.249\linewidth]{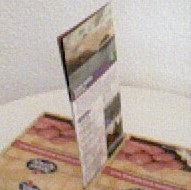}&\includegraphics[trim={0 0 0 0},clip,width=0.249\linewidth]{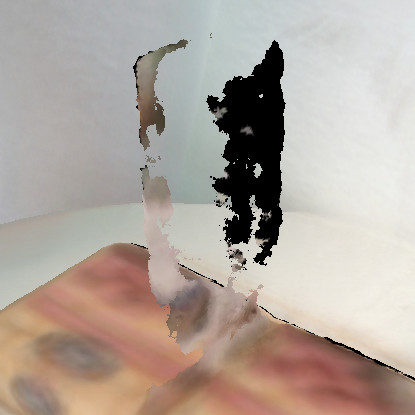}&\includegraphics[trim={0 0 0 0},clip,width=0.249\linewidth]{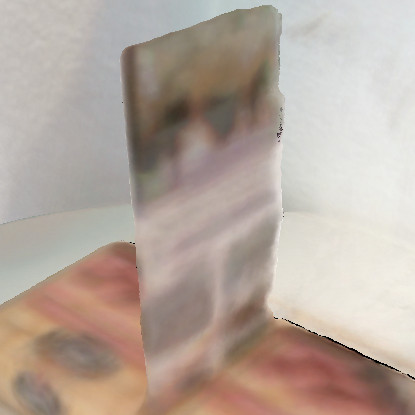}&\includegraphics[trim={0 0 0 0},clip,width=0.249\linewidth]{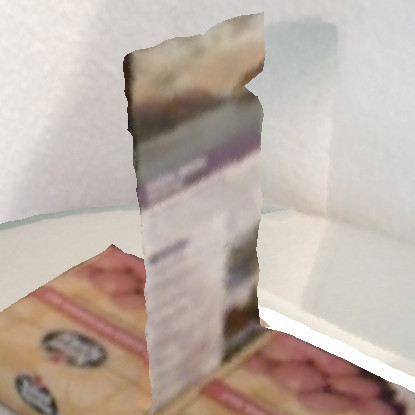}\\[0.3em]\footnotesize Photo & \footnotesize  \cite{infinitam2015ismar} (interim) & \footnotesize  \cite{infinitam2015ismar} & \footnotesize Ours 
\end{tabular}
\end{center}
\caption{Thin object reconstruction.
Top: front side, bottom: back side.
While ours reconstructs both sides successfully, InfiniTAM \cite{infinitam2015ismar} (using a voxel size of $1mm$) deletes one side before reconstructing the other (see the interim step shown) and finally reconstructs a mixture of the texture of both sides.}
\label{fig:thin_object}
\end{figure}

\PAR{Reconstructing thin objects.}
Another benefit of our method is its ability to reconstruct thin objects.
This is possible since it does not require surfaces to enclose a volume, and surfels with opposing normals do not interfere.
Thin objects are very hard to reconstruct correctly with volume-based methods, as their space discretization must be able to represent the object and the camera pose must be accurate enough to retain its volume.
We demonstrate this on the example of a flyer in Fig.~\ref{fig:thin_object}.

\begin{figure}[tb!]
\begin{center}
\renewcommand{\tabcolsep}{0px}
\renewcommand{\arraystretch}{0}
\begin{tabular}{ccc}
\includegraphics[trim={0 0 0 0},clip,width=0.25\linewidth]{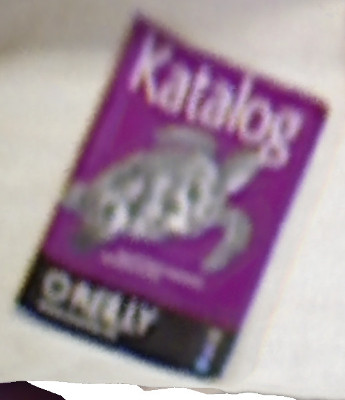}&\includegraphics[trim={0 0 0 0},clip,width=0.25\linewidth]{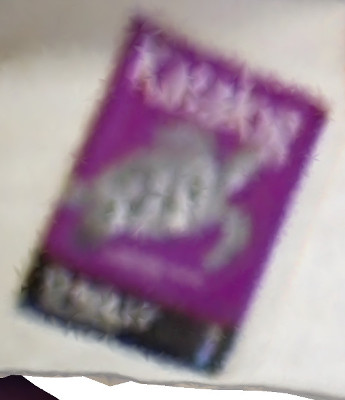}&\includegraphics[trim={0 0 0 0},clip,width=0.25\linewidth]{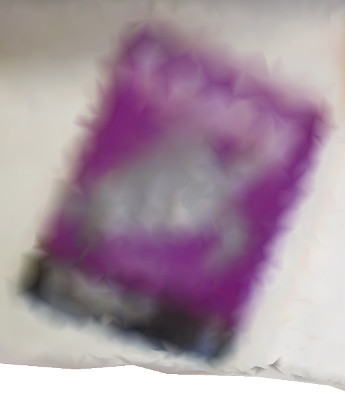}\\%
\includegraphics[trim={0 0 0 0},clip,width=0.25\linewidth]{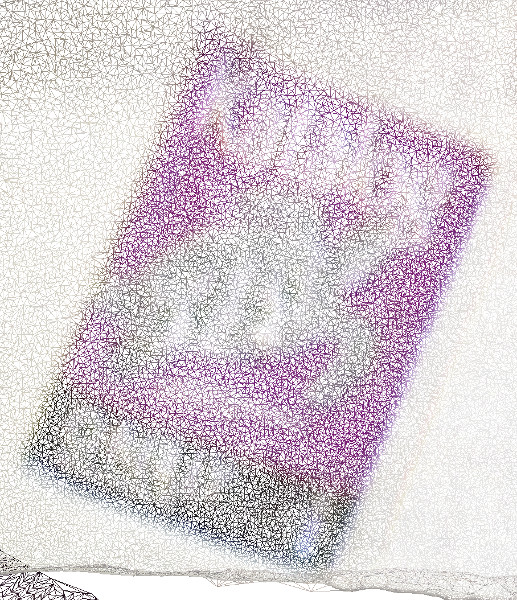}&\includegraphics[trim={0 0 0 0},clip,width=0.25\linewidth]{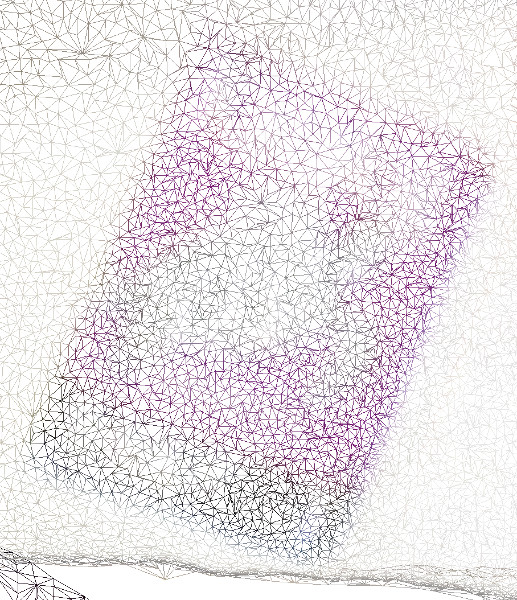}&\includegraphics[trim={0 0 0 0},clip,width=0.25\linewidth]{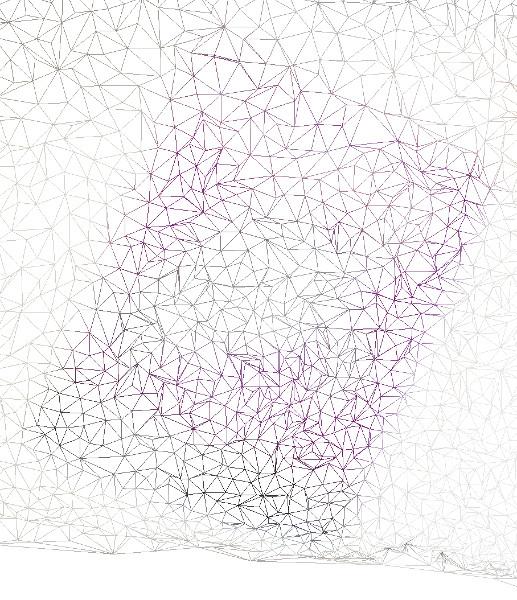}\end{tabular}
\end{center}
\caption{From left to right: reconstruction with full, half, and quarter resolution input.
With smaller input images, the mesh and color resolution decrease.}\label{fig:downsampling_texture_detail}
\end{figure}

\PAR{Impact of the input resolution.}
We downsample the input images to test the effect of using lower resolution input.
The original image size is $640{\times}480$ and the downsampled sizes are $320{\times}240$ and $160{\times}120$.
Since our outlier filtering (\cf paragraph on depth image preprocessing) depends on the image resolution, we slightly adapt it for the lower resolutions:
We do not remove image pixels which are within 2 pixels of a pixel without depth value.
This helps avoid removing too much geometry next to depth discontinuities at lower resolutions.
However, the gradient and radius calculation still require all neighbor pixels to be valid. 
Due to the resulting filtering, this leads us to remove more surface area at low resolutions compared to the full resolution.

\begin{figure*}[t]
\begin{center}
\renewcommand{\tabcolsep}{0px}
\renewcommand{\arraystretch}{0}
\begin{tabular}{ccc}
\includegraphics[trim={0 0 0 0},clip,width=0.28\linewidth]{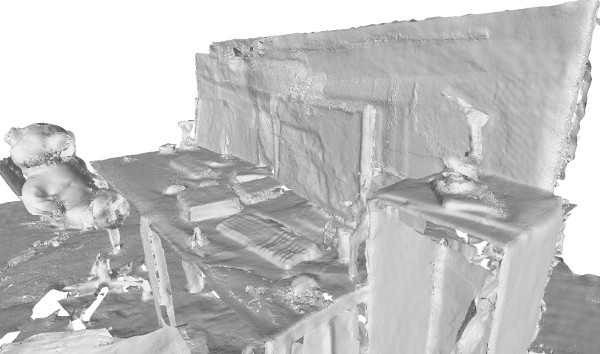}&\includegraphics[trim={0 0 0 0},clip,width=0.28\linewidth]{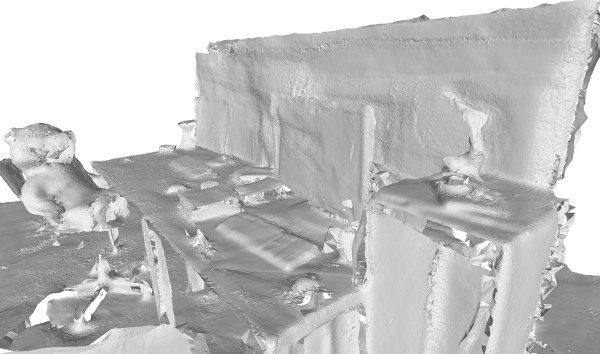}&\includegraphics[trim={0 0 0 0},clip,width=0.28\linewidth]{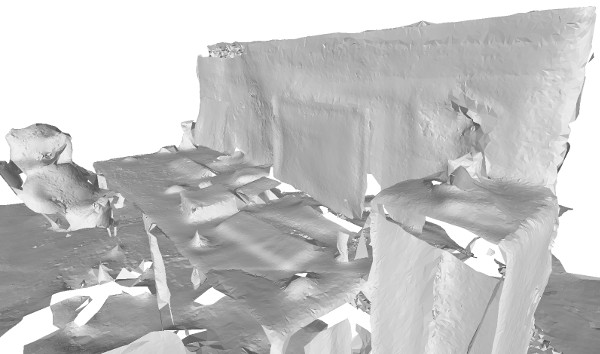}\end{tabular}
\end{center}
  \caption{From left to right: reconstruction with full, half, and quarter resolution input.
The geometry becomes less detailed as the resolution of the input images decreases. The mesh also becomes smoother due to stronger averaging of the sensor noise in the downsampling step and due to the denoising affecting larger areas, as well as the fact that small pose errors matter less at lower resolution.}
  \label{fig:downsampling_geometry_detail}
\end{figure*}

Fig.~\ref{fig:downsampling_texture_detail} shows a close-up view on how the mesh resolution, and thus the colors, change with the input resolution. 
As expected, using a lower input resolution leads to both a lower mesh resolution and a lower texture resolution. 

Fig.~\ref{fig:downsampling_geometry_detail} shows the changes in mesh geometry on a larger scale.
We note that the level of detail of the mesh degrades gracefully with the resolution of the RGB-D input images. 
However, lower resolution input leads to larger holes in the mesh, which we believe is partly due to the filtering when computing gradients and radii (mentioned above).

\subsection{Performance}
\label{sec:evaluation:performance}
\begin{figure}[t]
\begin{center}
\includegraphics[trim={0 0 0 0},clip,width=0.7\linewidth]{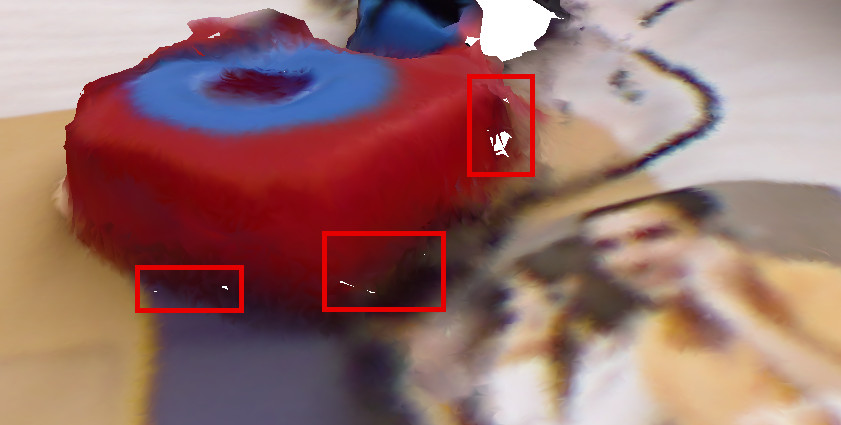}\end{center}
\caption{Example of tiny holes (highlighted in red) in a close-up of the reconstruction of fr1/desk \cite{sturm12iros} which is also shown in Fig.~\ref{fig:qualitative_reconstructions}.
The holes are created since the reconstructed surfels are too noisy for meshing in these areas.
While they are usually tiny, they can be quite noticeable if the mesh is rendered without anti-aliasing.}
\label{fig:limitation_holes}
\end{figure}
\begin{figure*}[t]
\begin{center}
\begin{tabular}{c}
\textbf{Original-size images ($\mathbf{640{\times}480}$ pixels)}\\
\includegraphics[trim={0 9pt 0 6pt},clip,width=0.9\linewidth]{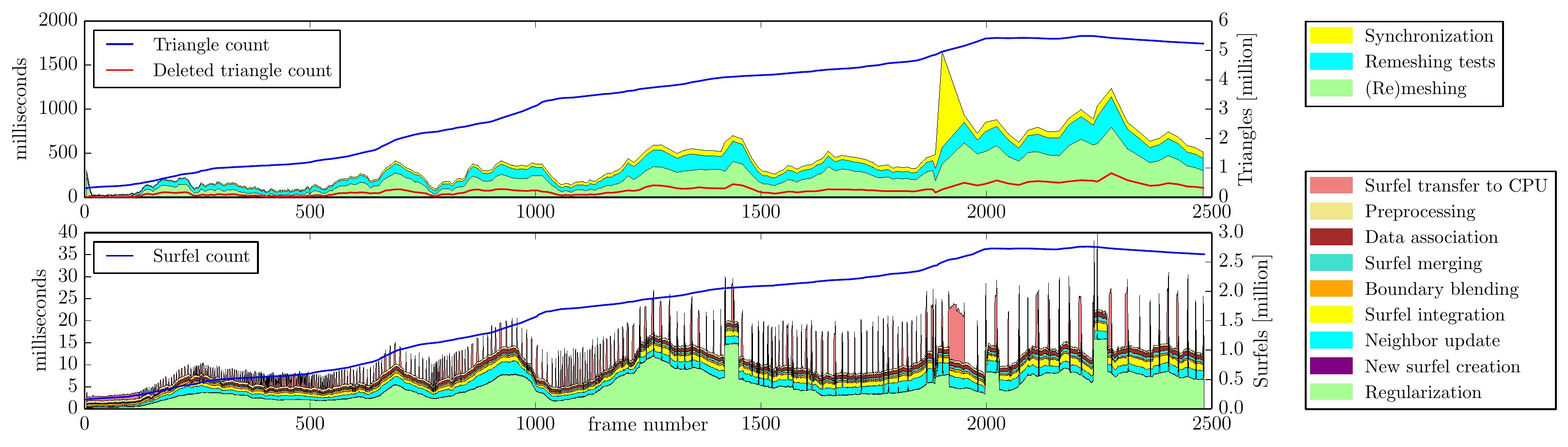}\\[0.5em]\hline~\\
\textbf{Half-size images ($\mathbf{320{\times}240}$ pixels)}\\
\includegraphics[trim={0 9pt 0 6pt},clip,width=0.9\linewidth]{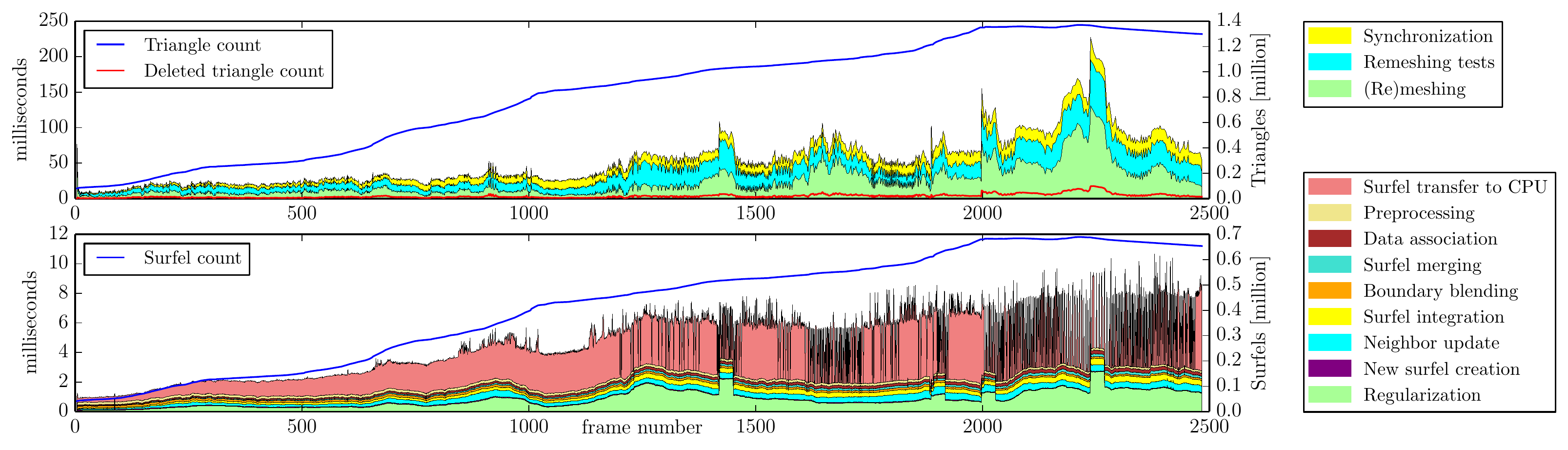}\\[0.5em]\hline~\\
\textbf{Quarter-size images ($\mathbf{160{\times}120}$ pixels)}\\
\includegraphics[trim={0 9pt 0 6pt},clip,width=0.9\linewidth]{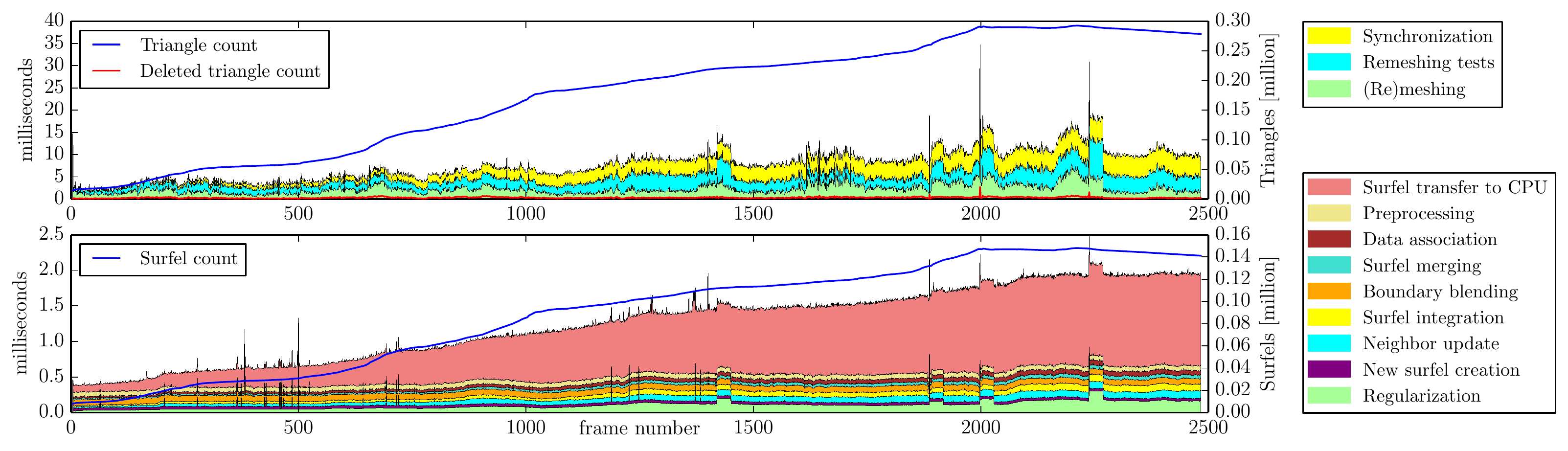}
\end{tabular}
\end{center}
\caption{Performance of (re)meshing (top of each plot pair) and surfel reconstruction (bottom of each plot pair) on the fr3/long\_office\_household dataset~\cite{sturm12iros}, for full input size (top), half size (middle), and quarter size (bottom).
The discontinuities in the surfel denoising performance are caused by re-activation of old surfels due to loop closures. 
These loop closures are also responsible for the increase in (re)meshing time by the end of the sequence.
Please notice the different scales.
}
\label{fig:performance}
\end{figure*}
Fig.~\ref{fig:performance} analyzes the performance of our approach on the dataset from Fig.~\ref{fig:teaser} depending on the image size. We optimized our (re)meshing implementation, but did not optimize surfel reconstruction and denoising,
which run in real-time for almost all frames using the original image resolution of $640{\times}480$. Due to the strong smoothing during loop closures, they cause short disruptions. 
For example, handling a loop closure for 3.1 million surfels takes ca.~$680 ms$.
Since loop closures are disruptive events in any case, we did not perceive this as an issue. 
In addition, the number of smoothing iterations could be reduced for higher speed. On average, remeshing deletes $6\%$ of the scene's triangles per iteration, and the average time per iteration is $212 ms$. 
Notice that since this article focuses on 3D reconstruction, the cost of external SLAM is not included in this analysis.

\section{Conclusion \& Discussion}
\label{sec:conclusion}

\PAR{Summary.}
We presented the first surfel-based approach for live mesh reconstruction from RGB-D video.
Compared to surfel reconstruction only, our approach creates a triangle mesh, which is useful for many applications.
Compared to volumetric methods, ours in particular handles loop closures with little effort, reconstructs high-resolution colors due to its ability to adapt the resolution of the 3D mesh to the input, and can reconstruct thin objects.
The approach can therefore provide a dense surface representation during SLAM. 
To the best of our knowledge, ours is the first method providing all these advantages during online 3D reconstruction.

\PAR{Limitations and future work.}
The Marching Cubes algorithm used by voxel-based methods creates manifold meshes.  In contrast, our meshes are not guaranteed to be manifold. As shown in Fig.~\ref{fig:limitation_holes}, our meshes can also contain holes if the smoothing is insufficient.  Future work could aim to resolve this by using temporary volumes for meshing, similar to \cite{ladicky2017point}, which might however also remove the ability to reconstruct thin objects. Still, as our evaluations show, non-manifoldness does not occur often.
In addition, improvements to depth cameras would also be expected to lead to higher-quality depth images and thus less holes. 

For loop closures inducing large deformations, the approach we adopted \cite{whelan2015elasticfusion} may rip apart surfaces.  In particular, if a surface is observed for a longer time, close-by surfels may be assigned to unrelated deformation nodes.
As future work, we believe that replacing the node assignment with a similar method from \cite{weise2011online}, which takes into account which nodes have been observed together, may help mitigating this.
Additional future work may be to transfer the boundary blending (which is independent from our surfel-based approach) or regularization also to volumetric approaches.

\PAR{Integration with SLAM.}
One of the motivations for this work was to enable surfel-based SLAM approaches to re-use their 3D representation for dense mesh reconstruction.
This avoids a duplication of effort compared to \eg, using a voxel-based reconstruction system in addition.
However, the most important requirements posed by 3D reconstruction are potentially different from those posed by SLAM.
For example, a 3D reconstruction is supposed to be complete and visually pleasing, while for SLAM, accuracy is of high importance instead.
The accuracy requirement of SLAM could conflict with the smoothing requirement of reconstruction, and the completeness requirement of reconstruction could mean that the SLAM system cannot handle the amount of data.
Another advantage of not integrating SLAM and reconstruction tightly is that this allows to easily replace one of the components.
Thus, the best way to combine SLAM and reconstruction is still an open question.
Since SurfelMeshing is a step towards enabling a tight integration between surfel-based SLAM and reconstruction, it might help answering this.
At the same time, the approach could be easily combined with different SLAM systems as long as its surfels can be updated based on the SLAM state, \eg, if surfel position updates are tied to updates in keyframe poses.

\ifCLASSOPTIONcompsoc
    \section*{Acknowledgments}
\else
    \section*{Acknowledgment}
\fi

Thomas Sch\"ops was supported by a Google PhD Fellowship.

\bibliographystyle{IEEEtran}
\bibliography{egbib}

\begin{IEEEbiography}[{\includegraphics[width=1in,height=1.25in,clip,keepaspectratio]{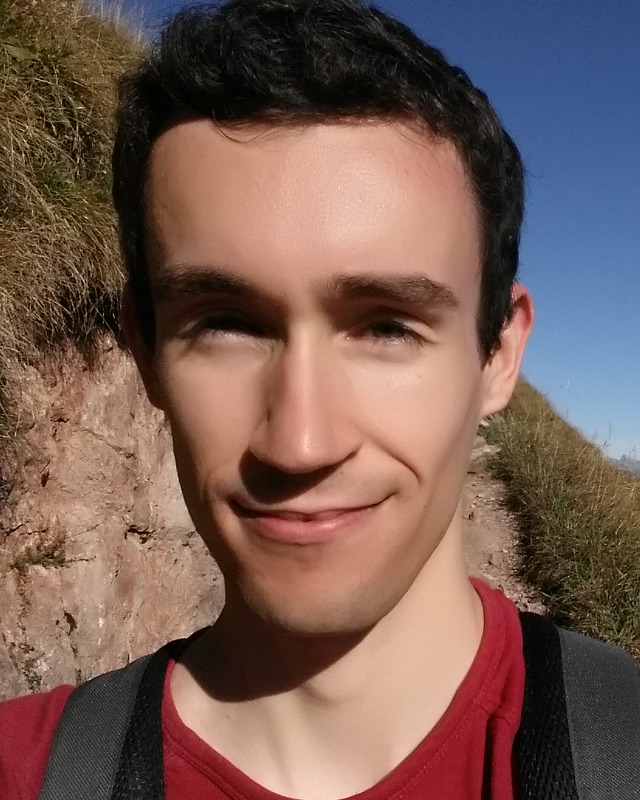}}]{Thomas Sch\"ops}
received the MS degree in Informatics from TU Munich in 2014. He is currently working towards the PhD degree in the Computer Vision and Geometry Group at ETH Zurich under the supervision of Marc Pollefeys. His research interests are dense 3D reconstruction and SLAM.
\end{IEEEbiography}

\begin{IEEEbiography}
[{\includegraphics[width=1in,height=1.25in,clip,keepaspectratio]{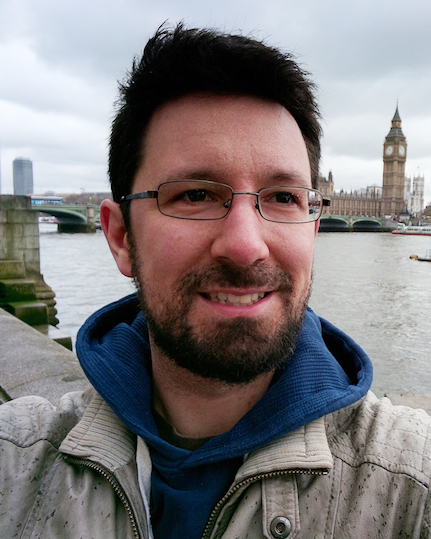}}]{Torsten Sattler}
received his PhD degree from RWTH Aachen University in 2014. He was a postdoctoral and senior researcher at ETH Zurich in the Computer Vision and Geometry Group, where he was Marc Pollefeys' deputy from 2016 to 2018. He is currently an associate professor at Chalmers University of Technology. His research interests center around visual localization and 3D mapping and include local features, camera pose estimation, SLAM, dense 3D reconstruction, image retrieval, and semantic scene understanding. He has organized workshops and tutorials on these topics at ECCV, ICCV, and CVPR.
\end{IEEEbiography}

\begin{IEEEbiography}
[{\includegraphics[width=1in,height=1.25in,clip,keepaspectratio]{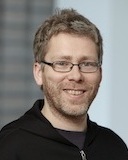}}]{Marc Pollefeys}
is a Professor of Computer Science at ETH Zurich and Director of Science at Microsoft working on HoloLens and Mixed Reality.  He is best known for his work in 3D computer vision, having been the first to develop a software pipeline to automatically turn photographs into 3D models, but also works on robotics, graphics and machine learning problems.  Other noteworthy projects he worked on with collaborators at UNC Chapel Hill and ETH Zurich are real-time 3D scanning with mobile devices, a real-time pipeline for 3D reconstruction of cities from vehicle mounted-cameras, camera-based self-driving cars and the first fully autonomous vision-based drone.  Most recently his academic research has focused on combining 3D reconstruction with semantic scene understanding.  He received a master of science in electrical engineering and a PhD in computer vision from the KU Leuven in Belgium in 1994 and 1999 respectively.  He became an assistant professor at the University of North Carolina in Chapel Hill in 2002 and joined ETH Zurich as a full professor in 2007. 
\end{IEEEbiography}

\end{document}